\documentclass{article}

\usepackage{wrapfig}

\usepackage[accepted]{icml2025}
\usepackage{microtype}
\usepackage{graphicx}
\usepackage{subfigure}
\usepackage{booktabs} %
\usepackage{arydshln} %
\usepackage{tikz}
\usetikzlibrary{bayesnet} %
\usetikzlibrary{arrows}
\usepackage{amsmath}
\usepackage{amssymb}
\usepackage{mathtools}
\usepackage{amsthm}

\usepackage{enumitem,multirow,adjustbox}

\usepackage{hyperref}
\usepackage{url}
\usepackage{xcolor}		%
\definecolor{darkblue}{rgb}{0, 0, 0.5}
\definecolor{beaublue}{rgb}{0.74, 0.83, 0.9}
\definecolor{gainsboro}{rgb}{0.86, 0.86, 0.86}
\definecolor{kleinblue}{rgb}{0,0.18,0.65}
\hypersetup{colorlinks=true,citecolor=kleinblue, linkcolor=kleinblue, urlcolor=kleinblue}

\usepackage{pifont}

\usepackage{amsmath,amsfonts,bm}

\def\eqref#1{equation~\ref{#1}}

\def\1{\bm{1}}

\def\rvtheta{{\mathbf{\theta}}}

\def\va{{\bm{a}}}

\def\vc{{\bm{c}}}

\def\ve{{\bm{e}}}

\def\vh{{\bm{h}}}

\def\vp{{\bm{p}}}
\def\vq{{\bm{q}}}

\def\vv{{\bm{v}}}

\def\vx{{\bm{x}}}

\DeclareMathAlphabet{\mathsfit}{\encodingdefault}{\sfdefault}{m}{sl}
\SetMathAlphabet{\mathsfit}{bold}{\encodingdefault}{\sfdefault}{bx}{n}

\def\gD{{\mathcal{D}}}
\def\gE{{\mathcal{E}}}

\def\gG{{\mathcal{G}}}

\def\gL{{\mathcal{L}}}
\def\gM{{\mathcal{M}}}
\def\gN{{\mathcal{N}}}

\def\gV{{\mathcal{V}}}

\def\gZ{{\mathcal{Z}}}

\newcommand{\R}{\mathbb{R}}

\DeclareMathOperator*{\argmin}{arg\,min}

\usepackage{enumitem,algorithm,algorithmic}

\usepackage{xspace}

\makeatletter
\newcommand*\rel@kern[1]{\kern#1\dimexpr\macc@kerna}
\newcommand*\widebar[1]{%
  \begingroup
  \def\mathaccent##1##2{%
    \rel@kern{0.8}%
    \overline{\rel@kern{-0.8}\macc@nucleus\rel@kern{0.2}}%
    \rel@kern{-0.2}%
  }%
  \macc@depth\@ne
  \let\math@bgroup\@empty \let\math@egroup\macc@set@skewchar
  \mathsurround\z@ \frozen@everymath{\mathgroup\macc@group\relax}%
  \macc@set@skewchar\relax
  \let\mathaccentV\macc@nested@a
  \macc@nested@a\relax111{#1}%
  \endgroup
}
\makeatother

\newenvironment{myquotation}{\setlength{\leftmargini}{0em}\quotation}{\endquotation}
\usepackage[hyperpageref]{backref}

\renewcommand*{\backrefalt}[4]{
  \ifcase #1 \relax
  \or
    (Cited on page #2)
  \else
    (Cited on pages #2)
  \fi
}

\definecolor{Gray}{gray}{0.9}

\graphicspath{{./fig/}}

\usepackage{colortbl}
\definecolor{aliceblue}{RGB}{178, 217, 245}

\definecolor{babyblue}{RGB}{217, 239, 251}

\definecolor{babypink}{RGB}{251, 231, 230}
\definecolor{mygreen}{HTML}{3cb44b}
\definecolor{purple}{HTML}{7D2882}
\definecolor{darkred}{HTML}{B22222}

\usepackage{etoc}
\etocdepthtag.toc{mtchapter}
\etocsettagdepth{mtchapter}{subsection}
\etocsettagdepth{mtappendix}{none}

\newcommand{\ours}[0]{\texttt{HIGHT}\xspace}
\newcommand{\hdata}[0]{\texttt{HiPubChem}\xspace}
\newcommand{\hbench}[0]{\texttt{MotifHallu}\xspace}
\newcommand{\ourst}[0]{\text{HIGHT}\xspace}	%
\newcommand{\oursfull}[0]{\textbf{HI}erarchical \textbf{G}rap\textbf{H} \textbf{T}okenization\xspace}
\newcommand{\molebert}[0]{\texttt{Mole-BERT}\xspace}
\newcommand{\chebi}[0]{\texttt{ChEBI-20}\xspace}
\newcommand{\molins}[0]{\texttt{Mol-Instructions}\xspace}
\newcommand{\molnet}[0]{\texttt{MoleculeNet}\xspace}

\icmltitlerunning{Hierarchical Graph Tokenization for Molecule-Language Alignment}

\begin{document}
\twocolumn[
\icmltitle{Hierarchical Graph Tokenization for Molecule-Language Alignment}

\begin{icmlauthorlist}
\icmlauthor{Yongqiang Chen}{mbz,cmu,cuhk}
\icmlauthor{Quanming Yao}{thu}
\icmlauthor{Juzheng Zhang}{umd}
\icmlauthor{James Cheng}{cuhk}
\icmlauthor{Yatao Bian}{nus}
\end{icmlauthorlist}

\icmlaffiliation{nus}{Department of Computer Science, National University of Singapore}
\icmlaffiliation{cuhk}{The Chinese University of Hong Kong}
\icmlaffiliation{thu}{Tsinghua University}
\icmlaffiliation{umd}{University of Maryland, College Park}
\icmlaffiliation{mbz}{MBZUAI}
\icmlaffiliation{cmu}{Carnegie Mellon University}

\icmlcorrespondingauthor{Yatao Bian}{bianyt@comp.nus.edu.sg}

\icmlkeywords{Large Language Models, Graph Neural Networks, Molecule Understanding, Alignment, Tokenization}

\vskip 0.3in
]

\printAffiliationsAndNotice{Most of the works were done when Yongqiang Chen was a PhD student at CUHK.} %

\begin{abstract}
    Recently, there has been a surge of interest in extending the success of large language models (LLMs) from texts to molecules.
    Most existing approaches adopt a graph neural network to represent a molecule as a series of node tokens for molecule-language alignment,
    which, however, have overlooked the inherent hierarchical structures in molecules. 
    Notably,  higher-order molecular structures contain rich semantics of functional groups, which encode crucial biochemical functionalities of the molecules.
    We show that neglecting the hierarchical information in tokenization will lead to subpar molecule-language alignment and severe hallucination. 
    To address this limitation, we propose \oursfull (\ours).
    \ours employs a hierarchical graph tokenizer that encodes the hierarchy of atom, motif, and molecular levels of informative tokens to improve the molecular perception of LLMs.
    \ours also adopts an augmented instruction tuning dataset, enriched with the hierarchical graph information, to further enhance the molecule-language alignment.
    Extensive experiments on $14$ real-world benchmarks verify the effectiveness of \ours in reducing hallucination by $40$\%, and significant improvements in various molecule-language downstream tasks. 
    The project is available at \url{https://higraphllm.github.io/}.
\end{abstract}

\section{Introduction}
Large language models (LLMs) have demonstrated impressive capabilities in understanding and processing natural languages~\citep{gpt2,chatgpt,llama,spark_AGI}.
Recently, there has been a surge of interest in extending the capabilities of LLMs to graph-structured data~\citep{Jin2023LargeLMGraph,Li2023GraphMeetLLM,Wei2024TowardsVGraph,Mao2024GraphFM,Fan2024GraphML}, particularly molecular graphs~\citep{zhao2023gimlet,cao2023instructmol}. 
Inspired by the success of large vision-language models~\citep{zhang2024vision,liu2023llava}, recent efforts in developing large graph-language models (LGLMs) typically adopt a graph neural network (GNN)~\citep{gin} to tokenize molecules as a series of node embeddings (or node tokens), and then leverage an adapter such as a Multi-layer perceptron (MLP) or a Q-former~\citep{blip2} to transform the node tokens into those compatible with LLMs~\citep{Fan2024GraphML}. 
To bridge the gap between the graph and language modalities, LGLMs will undergo a molecule-language instruction tuning with the molecular graph and the corresponding captions describing the molecules~\citep{Jin2023LargeLMGraph,Li2023GraphMeetLLM,Fan2024GraphML}.

\begin{figure*}[t]
    \centering
    \subfigure[Overview of the \ourst framework.]{
        \includegraphics[width=0.6\textwidth]{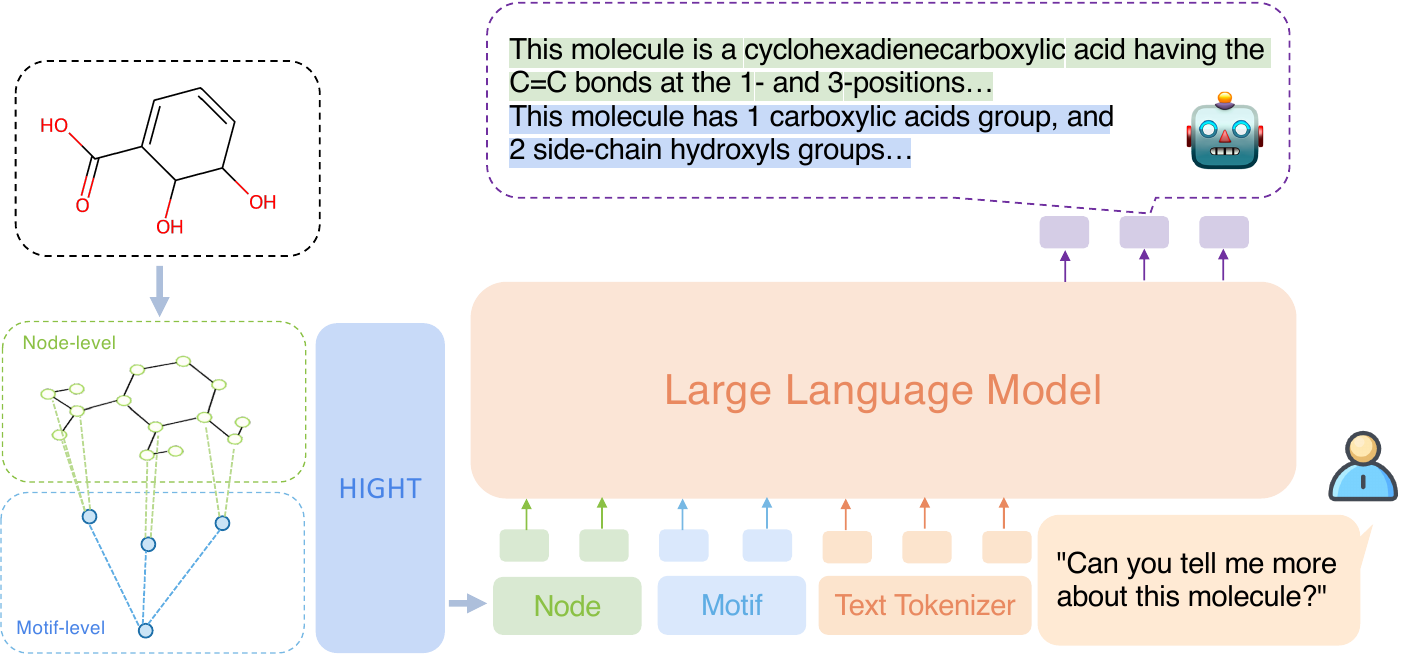}
    }
    \subfigure[Summary of performance.]{
        \includegraphics[width=0.35\textwidth]{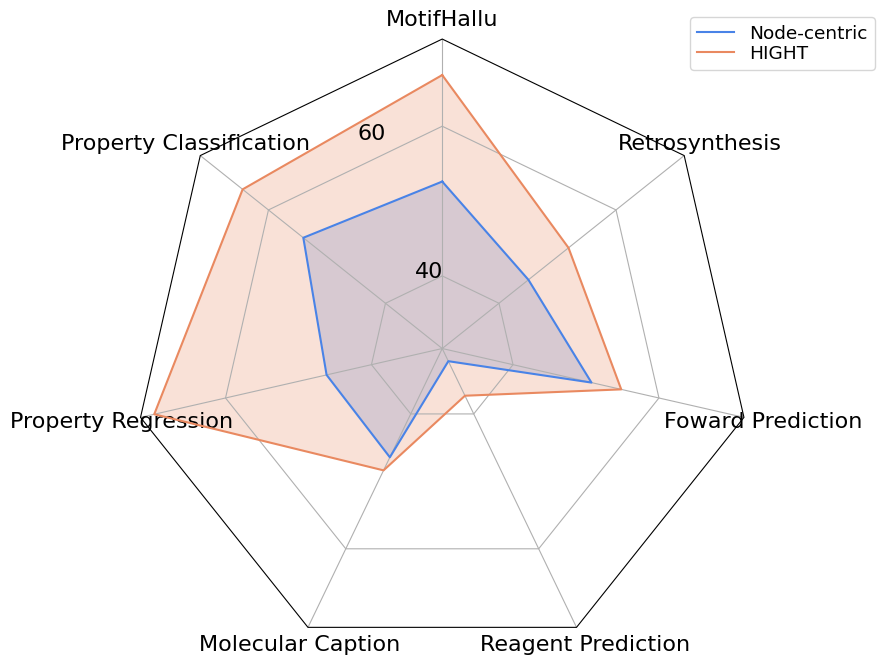}
    }
    \caption{\textit{(a)} \textbf{Illustration of \ourst}: Given a molecule (i.e., PubChem ID 3, \textit{5,6-Dihydroxycyclohexa-1,3-diene-1-carboxylic acid}), \ours detects the motifs and incorporates the ``supernodes'' for each motif (The whole graph is also considered as a ``super motif''.). Then, \ours tokenizes the molecule into both node-level (i.e., atoms) and motif-level (i.e., functional groups) tokens. The hierarchical view enables LLMs to align the molecular structures and the language descriptions of the molecule better. \textit{(b)} \textbf{Performance Overview}: \ours significantly reduces the hallucination of LGLMs and improves the downstream performance across various molecule-centric tasks. Due to the heterogeneity of the evaluation metrics in each task, we perform some transformations on the numerical values. In MotifHallu, we report the macro F1 scores. For Property Classification and Molecular Caption, we report the averaged scores of all the subtasks or submetrics. For Property Regression, we normalize the values to the range between $1$ and $100$, i.e., for $a$, the reported number is $0.5/a$. For Chemical Reaction Prediction, we report the averaged values of BLEU, RDK, MACCS, and MORGAN.}
    \label{fig:motivation}
\end{figure*}

Despite recent progress, the tokenization in existing LGLMs neglects the \textit{essential hierarchical structures} inherent in molecular graphs. In particular, in molecular graphs, the high-order substructures, such as motifs or functional groups, encode rich semantics of the biochemical functionalities of the molecules~\citep{Milo2002NetworkMS,art_drug,zinc15}. For example, the presence of a hydroxide functional group (``-OH'') often indicates a higher water solubility.
Therefore, such substructural cues are essential for enabling LLMs to reason about the molecules in a chemically meaningful way.
However, existing LGLMs mostly tokenize molecules solely at the atom (node) level, and feed LLMs with only node-level tokens.
Consequently, it requires LLMs to implicitly infer the underlying substructures during the instruction tuning stage.
The absence of the critical substructures not only increases the unnecessary burdern on the LLMs, but also leads to misaligned representations and a higher likelihood of hallucinations in downstream tasks.
To quantify the issue, we introduce a diagnostic benchmark, called \hbench, which evaluates the perception ability of LGLMs about the existence of common functional groups. Surprisingly, we find that existing LGLMs often produce false-positive predictions (i.e., keep answering ``Yes'' for any functional groups), highlighting a critical limitation in current graph tokenization strategies (Sec.~\ref{sec:motif_hallu}). This observation motivates the following research question:
\begin{myquotation}\centering
    \textit{Is there a  feasible approach to integrate the intrinsic hierarchical molecular information into LLMs?}
\end{myquotation}
To tackle the problem, we propose a new molecule-language alignment strategy called \oursfull (\ours). As illustrated in Fig.~\ref{fig:motivation}, \ours adopts a hierarchical graph tokenizer and a hierarchical molecular instruction tuning dataset to facilitate a better alignment of molecule and language modalities. 
Specifically, inspired by the success of hierarchical GNNs in molecular representation learning~\citep{MGSSL,zang2023hierarchical,inae2023motifaware,2023fragmentbased}, \ours transforms the original molecular graph into a hierarchical graph with motif-level and molecule-level nodes added in. Then, \ours employs a Vector Quantized-Variational AutoEncoder (VQVAE) to obtain atom-level, motif-level, and molecule-level tokens separately with the self-supervised tasks~\citep{zang2023hierarchical}. 

In addition, to further encourage the encoding and alignment of hierarchical information, \ours augments the original molecular instruction tuning dataset with motif-level descriptions.
Our contributions can be summarized as follows:
\begin{itemize}[leftmargin=*]
    \item To the best of our knowledge, we are the \textit{first to incorporate the hierarchical graph information} into LGLMs, with the consideration of both the architecture-level and the instruction tuning data.
    \item To facilitate the molecule-language alignment study, we also propose the \textit{first hallucination benchmark} \hbench, synthesized through question-answering based on common functional groups.
    \item We conduct extensive experiments with $14$ real-world benchmarks. The results show that \ours significantly reduces the hallucination on \hbench by up to $40$\% and consistently improves the performances on downstream molecule-language tasks.
\end{itemize}
Hence, \ours together with \hbench and \hdata, lay the solid foundation for developing graph foundation models via graph-language alignment.

\section{Preliminaries}

\textbf{Large Graph-Language Models.}
As LLMs have demonstrated great capabilities across a wide range of natural language tasks, there has been an increasing interest in extending LLMs to broader applications where the text data are associated with the structure information (i.e., graphs)~\citep{Jin2023LargeLMGraph,Li2023GraphMeetLLM,Wei2024TowardsVGraph,Mao2024GraphFM,Fan2024GraphML}. 
A graph can be denoted as $\gG=(\gV,\gE)$ with a set of $n$ nodes $v\in\gV$ and a set of $m$ edges $(u,v)\in\gE$. Each node $u$ has node attributes as $\vx_u\in\R^d$ and each edge $(u,v)$ has edge attributes $e_{u,v}\in\R^{d_e}$.
A number of LGLMs have been developed to process graph-text associated data $\gD=\{\gG,\vc\}$, where $\vc=[c_1,...,c_{l_c}]$ is to the caption of the graph $\gG$. For node-centric tasks, $\vc_i$ will associate with the nodes~\citep{tang2023graphgpt}, while in this paper we focus on graph-centric tasks, i.e., molecules and molecular captions~\citep{liu2023molca}.
Usually, an $l$-layer GNN is employed to encode a graph as:
\begin{equation}
    \vh^{(l)}_u=\text{COM}(\vh^{(l-1)}_u,\text{AGG}(\{(\vh^{(l-1)}_u,\vh^{(l-1)}_v)|v\in\gN(u)\})),
\end{equation}
where $\vh^{(l)}_u\in\R^h$ refers to the node embedding of node $u$ after $l$ layers of GNN, $\text{AGG}(\cdot)$ is the aggregation function (e.g., mean) among the information from neighbors of node $u$, and $\text{COM}$ is the operator for combining information of node $u$ with its neighbors $\gN(u)$ (e.g., concatenation). Then, after $l$ message passing iterations, the graph-level embedding can be obtained as:
\begin{equation}
    \vh_\gG=\text{READOUT}\left(\{h^{(l)}_u|u\in\gV\}\right),
\end{equation}
where $\text{READOUT}(\cdot)$ is a pooling operator (e.g., mean pooling) among all the node embeddings.
With the representations of the nodes and graphs, LGLMs can fuse the graph and language information in various ways, such as transforming into natural languages describing the graphs~\citep{fatemi2024talk}, or neural prompts within the LLMs~\citep{GNP}. In addition, the embeddings can also be leveraged to post-process the LLM outputs~\citep{liu2024oneforall}.
Orthogonal to different fusion mechanisms, in this work, \textit{we focus on transforming graph embeddings into input tokens of LLMs}, which can be formulated as~\citep{tang2023graphgpt,Chen2024LLaGALL,liu2023molca,zhao2023gimlet,cao2023instructmol,li2024Mol3D}:
\begin{equation}
    p_\rvtheta(\va|\vq,\vh)=\text{$\prod$}_{i=1}^{l_a}p_\rvtheta(\va_i|\vq,f_n(\vh),\va_{<i}),
\end{equation}
where the LGLM is required to approximate $p_\rvtheta$ to output the desired answer $\va$ given the question $\vq$, and the graph tokens $\vh$ adapted with adapter $f_n:\R^h\rightarrow\R^{h_e}$ that projects the graph tokens to the embedding space of LLMs.
One could also incorporate the 1D information such as SMILES~\citep{smiles} into $\vq$ and $\va$ for alignment.

\textbf{Molecular Foundation Models.} There is a separate line of works aiming to develop language models for molecules and proteins -- the language of lives, from 1D sequences such as SMILES~\citep{Chemformer}, 2D molecular graphs~\citep{grover,molclr,zhang2024unimot}, 3D geometric conformations~\citep{graphmvp,zhou2023unimol}, to scientific text~\citep{scibert} and multimodal molecule-text data~\citep{liu2023moleculestm,luo2023molfm,Text+ChemT5,liu2024gitmol,MoMu,kv-plm,Srinivas2024CrossingNF}. The adopted backbones range from encoder-decoder architectures such as MolT5~\citep{MolT5} and Galactica~\citep{GalacticaAL}, to auto-regressive language modeling~\citep{luo2023biomedgpt,molxpt}.
Inspired by the success of large vision-language models~\citep{blip2,zhu2023minigpt,liu2023llava},
the community further seeks to develop molecular foundation models built upon existing molecular language models with more sophisticated graph information fusion modules. 
For example, \citet{liu2023molca,zhao2023gimlet} develop advanced cross-modal adapters and generalized position embeddings to promote better alignment based on encoder-decoder-based molecular language models. \citet{liang2023drugchat,cao2023instructmol,li2024Mol3D} develop cross-modal adapters for decoder only language models such as Llama~\citep{llama}. Orthogonal to the aforementioned works, we focus more on \textit{what information one shall extract from the molecules for the alignment}. We choose to build our methods upon decoder-only language models, with the hope of building a versatile agent that can perceive molecules beyond the language, image, and audio modalities~\citep{xi2023rise}.

In the meantime, existing works also try to enrich the molecule-language alignment with additional modalities, such as 2D~\citep{liu2023molca} and 3D~\citep{li2024Mol3D} information. In contrast, we focus on the intrinsic hierarchical information of the molecules, such as motifs.

\textbf{Hierarchical Graph Representation Learning.}
The hierarchical nature has been widely incorporated in learning high-quality graph representations~\citep{diffpool}. Especially in molecular graphs, the high-order structural information naturally captures the existence of motifs and functional groups. Therefore, the hierarchy of atom-motif-molecule has been widely applied in self-supervised molecular representation learning~\citep{MGSSL,zang2023hierarchical,inae2023motifaware,2023fragmentbased}. Nevertheless, how to properly incorporate the hierarchical information in molecular instruction tuning of LGLMs remains unclear.

In addition, concurrent works by~\citet{park2024llamo} and~\citet{Hu2024ExploringHM}  explored incorporating hierarchical graph information into LLMs. Nevertheless, they mostly focus on the architecture-level incorporation, while we show that it is  crucial to integrate  the hierarchical information in the instruction tuning data. More importantly, we highlight the consequences of inadequate alignment due to the lack of hierarchical information, i.e., hallucination, and demonstrate the usefulness of the hierarchical information in a wide range of downstream tasks.

\section{Graph Tokenization in LGLMs}
\label{sec:tokenization}
In this section, we analyze the limitations of node-centric tokenization, which is widely adopted in existing LGLMs.

\subsection{Node-Centric Tokenization}
\label{sec:node_token}
Specifically, most existing LGLMs directly take the node tokens from GNNs as inputs to LLMs~\citep{cao2023instructmol}:
\begin{equation}\label{eq:node_token}
    p_\rvtheta(\va|\vq,\vh)=\text{$\prod$}_{i=1}^{l_a}p_\rvtheta(\va_i|\vq,f_n(\vh_{1}),...,f_n(\vh_{n}),\va_{<i}),
\end{equation}
where $\vh_{1},...,\vh_{n}$ are node embeddings from a GNN typically pretrained through self-supervised learning on large-scale molecular datasets such as ZINC250k~\citep{zinc15}, $f_n$ is the adapter to project the node tokens to the LLM tokens.
There are various options to tokenize a molecule~\citep{liu2023rethinkingTokenizer}. In this work, we consider a state-of-the-art tokenizer~\citep{molebert} that pretrains a VQVAE~\citep{vqvae} with masked atoms modeling and constructs a codebook $\gZ$ to discretize atoms:
$z_u=\text{$\argmin$}_{i}||\vh_u-\ve_i||_2$,
where $z_u\in\gZ$ is the quantized index of atom $u$, and $\ve_v$ is the codebook embedding of the $i$-th entry. The codebook is trained through a reconstruction loss with respect to some attribute $\vv_i$ of atom $i$:
\begin{equation}
    \begin{aligned}
        \gL_{r}=&\frac{1}{n}\sum_{i=1}^n(1-\frac{\vv_i^T\hat{\vv_i}}{||\vv_i||\cdot||\hat{\vv_i}||})^\gamma+\frac{1}{n}\sum_{i=1}^n||\text{sg}[\vh_i]-\ve_{z_i}||_2^2\\
    &+\frac{\beta}{2}\sum_{i=1}^n||\text{sg}[\ve_{z_i}]-\vh_i||_2^2,
    \end{aligned}
\end{equation}
where $\text{sg}[\cdot]$ is the stop-gradient operator in straight-through estimator~\citep{ST_estimator}, $\hat{\vv_i}$ is the reconstructed attribute of atom $i$ with a decoder, and $\beta$ is a hyperparamter. In \molebert, the attribute is simply the type of atom. \molebert also manually partitions the codebook into groups of common atoms such as carbon, nitrogen, and oxygen to avoid codebook conflicts~\citep{molebert}.

\begin{figure*}[t]
    \centering
    \subfigure[Node-centric tokenization.]{
        \includegraphics[width=0.45\textwidth]{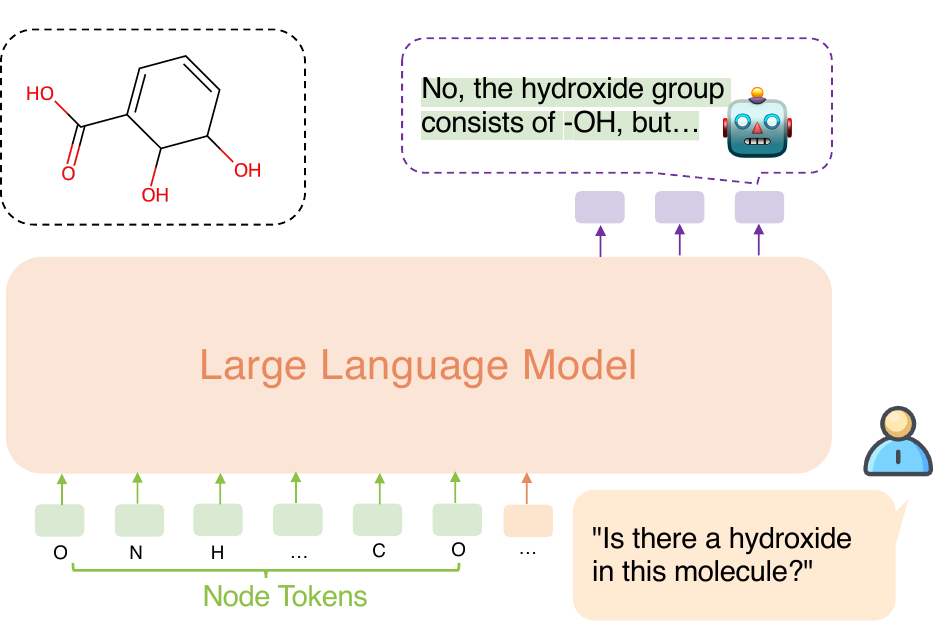}
    }
    \subfigure[\ourst tokenization.]{
        \includegraphics[width=0.45\textwidth]{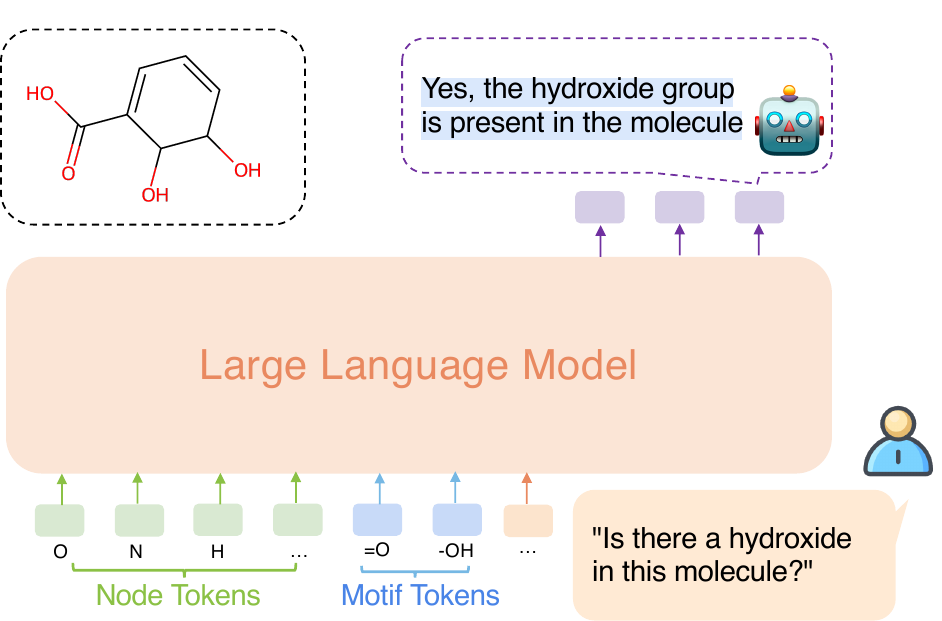}
    }
    \vspace{-0.1in}
    \caption{Illustration of hallucination caused by node-centric tokenization. With only node-level tokens, LLMs have to relate the nodes within a specific functional group to align useful molecular structures with the corresponding language descriptions. Yet, due to the arbitrary order of atoms and position biases in LLMs, it is hard to recognize each functional group, leading to severe hallucinations.}
    \label{fig:motif_hallu}
\end{figure*}

Intuitively, the trained atom tokens encode some contextual information, such as the neighbors of the atoms. 
However, node-centric tokenization makes the molecule-language alignment more challenging, as LLMs have to additionally relate the multiple nodes to align the corresponding texts during the instruction tuning process. 
Specifically, in molecules, motifs or functional groups usually capture rich semantics, and often share many common atoms such as carbon, nitrogen, and oxygen~\citep{art_drug}. 
As shown in Fig.~\ref{fig:motif_hallu}, both the carboxylic acid (``R-COOH'') and the hydroperoxide (``R-OOH'') functional groups all contain two oxygen atoms and a hydrogen atom. For a molecule with hydroperoxide attached to a scaffold with carbon atoms, it would be hard for LLMs to distinguish which functional group is present in the molecule.
Furthermore, due to the loss of positional information in the node-centric tokenization~\citep{liang2023drugchat,cao2023instructmol}, the limited expressivity of GNNs~\citep{gin} and the positional biases of auto-regressive LLMs~\citep{order}, it is more challenging for LLMs to relate the desired nodes in a motif, which will lead to subpar molecule-language alignment.

\subsection{Motif Hallucination}
\label{sec:motif_hallu}
To understand the issue of node-centric tokenization more clearly, we construct a simple benchmark called \hbench, to concretize the hallucination of common functional groups by LGLMs.
Specifically, we consider the $38$ common functional groups in RDKit\footnote{\url{https://github.com/rdkit/rdkit/blob/master/Data/FunctionalGroups.txt}} and leverage RDKit~\citep{rdkit} to detect the existence. We adopt $3,300$ molecules from \chebi~\citep{edwards2021text2mol} and query the existence of a functional group:
\begin{myquotation}\centering
    \texttt{Is there a <functional group name> in the molecule?}
\end{myquotation}
Then, we examine the outputs from LGLM meaning ``Yes'' or ``No''.
For each molecule, we construct questions with positive answers for all kinds of functional groups detected in the molecule, and questions with negative answers for randomly sampled $6$ functional groups from the remaining.
Hence \hbench consists of $23,924$ questions. While it is easy to scale up \hbench with more molecules and functional groups, we find that the current scale is already sufficient to demonstrate the issue (Table~\ref{tab:hbench}).

\section{Hierarchical Graph Tokenization}
\label{sec:hight_solution}
To improve the molecule-language alignment, we propose a new strategy called \oursfull (\ours), which contains a hierarchical graph tokenizer and a hierarchical molecular instruction tuning dataset to augment the inputs with hierarchical information.

\subsection{Hierarchical Graph Tokenizer}
\label{sec:hight_tokenizer}
Inspired by the success of hierarchical GNNs~\citep{MGSSL,zang2023hierarchical}, we transform the original molecular graph $\gG$ into a hierarchical graph $\gG'$ with motif-level and molecule-level nodes added in. Specifically, we leverage the Breaking of Retrosynthetically Interesting Chemical Substructures (BRICS) algorithm~\citep{BRICS}\footnote{Note that \ours  possesses a high degree of extensibility and can be augmented by incorporating advanced motif extraction techniques (such as \citep{MGSSL}).} to detect and inject a set of $k+1$ supernodes, denoted as $\gM=\{\gM^{(1)},...,\gM^{(k)},\gM^{(k+1)}\}$, with $k$ motifs and the original molecule $\gM^{(k+1)}=\gG$.
Furthermore, denoting the set of nodes and edges in $\gM^{(i)}$ as $\gV^{(i)}_m$ and $\gE^{(i)}_m$, respectively, we augment the original molecular graph $\gG$ as $\gG'$ with augmented nodes $\gV'$ and edges $\gE'$:
\begin{equation}
    \gV'=\gV\cup\{v^{(1)}_m,...,v^{(k+1)}_m\},\ 
    \gE'=\gE\cup(\cup_{i=1}^{k+1}\gE_{ma}^{(i)}),
\end{equation}
where $v^{(i)}_m$ is the motif super nodes added to the original molecule, and $\gE_{ma}^{(i)}=\cup_{u\in\gV_m^{(i)}}\{(u,v^{(i)}_m)\}$ are the augmented edges connecting to the motif super node from nodes within the corresponding motif.
We employ separate VQVAEs for atoms and motifs to learn meaningful code embeddings with several self-supervised learning tasks. The reconstructed attributes in Eq.~\ref{eq:node_token} include atom types at the atom-level and the number of atoms at the motif-level~\citep{zang2023hierarchical}.

Merely feeding the motif tokens with node tokens to LLMs still can not help distinguish the motifs from atoms properly, hence we propose to further attach positional encodings $\vp$ to all of the tokens. We choose to use Laplacian positional embeddings~\citep{dwivedi2020benchmarkgnns} while one could also adopt other variants~\citep{graph_transformer}. Since different types of tokens contain distinct semantics, we adopt separate adapters for different types of tokens. 
Denoting the motif tokens as $\vh_m^{(i)}$ for motif $\gM^{(i)}$, generation with \ours is:
\begin{equation}\label{eq:hight_token}
    \begin{aligned}
        p_\rvtheta(\va|\vq,\vh,\vh_m)=&\text{$\prod$}_{i=1}^{l_a}p_\rvtheta(\va_i|\vq,f_n(\vh_{1}),...,f_n(\vh_{n}),\\
        &f_m(\vh_{m}^{(1)}),...,f_g(\vh_m^{(k+1)}),\va_{<i}),
    \end{aligned}
\end{equation}
where $f_m(\cdot)$ and $f_g(\cdot)$ are the adapters for BRICS motifs and the original molecules, respectively.

\subsection{Hierarchical Graph Instruction Tuning Dataset}
\label{sec:hight_data}
Although \ours tokenizer properly extracts the hierarchical information from the input graph modality, it remains challenging to properly align the language to the corresponding molecular information, without the appearance of the respective captions in the texts.
For example, if the caption does not contain any information about the water solubility of the hydroxide functional group (``-OH''), LGLMs will never know that ``-OH'' motif corresponds to the water solubility of the molecule, despite that \ours tokenizer extracts the ``-OH'' token.
In fact, the commonly used molecular instruction tuning curated from PubChem~\citep{pubchem} in existing LGLMs~\citep{liu2023molca,cao2023instructmol,li2024Mol3D}, contains surprisingly little information about motifs.
Some samples are given in Appendix~\ref{appdx:hdata}.

To this end, we propose \hdata, which augments the molecular instruction tuning dataset with captions of the functional groups. 
We consider both the positive and negative appearances of motifs: For the positive case, we directly append the caption of all functional groups detected with RDKit.
We also include a brief introduction of the functional groups to provide fine-grained information for molecule-language alignment.
For the negative case, we randomly sample $k_\text{neg}$ motifs not appeared in the molecule to explicitly instruct LGLMs on the absence of the motifs.
Despite the simple augmentation strategy, we find that \hdata significantly reduces the hallucination issue and improves the alignment performance.

\subsection{Hierarchical Graph Instruction Tuning}
\label{sec:hight_sft}
We use a two-stage instruction tuning~\citep{cao2023instructmol}.

\textbf{Stage 1 Alignment Pretraining.} We curate a new molecule-text paired dataset from PubChem following the pipeline of~\citet{liu2023moleculestm}. We set the cutoff date by Jan. 2024, and filter out unmatched pairs and low-quality data, which results in $295$k molecule-text pairs. Furthermore, we construct the \hdata-$295$k dataset. The first stage mainly warms up the adapter to properly project the graph tokens with the LLM embedding space. To avoid feature distortion, both the LLM and the GNN encoder are frozen.

\textbf{Stage 2 Task-specific Instruction Tunning.}
With a properly trained adapter, we further leverage the task-specific instruction tuning datasets from \molnet~\citep{molnet}, \chebi~\citep{chembl}, and \molins~\citep{mol-ins}. More details are given in Appendix~\ref{appdx:dataset}. In Stage 2, we still keep the GNN encoder frozen, while tuning both the adapter and the LLM (with low-rank adaptation, i.e., LoRA~\citep{hu2022lora}).

\begin{table*}
\centering
\small
\caption{\footnotesize
Detailed results in motif hallucinations on \hbench. Due to the imbalance of samples from positive and negative classes, we incorporate diverse evaluation metrics to provide a detailed comparison between different methods in terms of hallucination.
}
\label{tab:hbench_full}
\scalebox{0.95}{
\begin{tabular}{lccccccc}
    \toprule
    \textsc{Method} &Macro F1 $\uparrow$&F1 (pos) $\uparrow$  &F1 (neg) $\uparrow$ &Micro F1 $\uparrow$ &AUROC $\uparrow$&Acc $\uparrow$  &Yes Ratio \\
    & & $4,124$ &$19,800$ && & & \\
    \midrule
    {GIMLET~\citep{zhao2023gimlet}}& 50.0 & 0.1 & \textbf{99.9} &0.05&49.9&\textbf{82.6}&0.2\\
    {Galactica-6.7B~\citep{GalacticaAL}}&56.6&17.5&95.7&12.9&50.7&77.6&8.5\\
    {InstructMol~\citep{cao2023instructmol}}&52.6&\textbf{95.7}&9.5&28.3&48.4&20.0&94.5\\
    \textbf{\ourst}&\textbf{66.8}&85.5&48.2&\textbf{29.7}&\textbf{53.2}&39.1&69.4\\
    \bottomrule
\end{tabular}
}
\end{table*}

\begin{table}
\centering
\small
\vspace{-0.1in}
\caption{\footnotesize
Results of motif hallucinations on \hbench.
}
\label{tab:hbench}
\scalebox{0.75}{
\begin{tabular}{lcccc}
    \toprule
    \textsc{Method} &F1 (pos) $\uparrow$  &F1 (neg) $\uparrow$ &Acc $\uparrow$  &Yes Ratio \\
    \midrule
    
    \rowcolor[rgb]{0.94, 0.97, 1.0} \multicolumn{5}{l}{\textit{Node-centric Tokenization}} \\
    {InstructMol-G} & 95.7 & 9.5 & 19.9 & 94.5\\
    {InstructMol-G} (LLama-2-7b-chat) & \textbf{99.6} & 2.8 & 18.3 & 98.7\\
    {InstructMol-GS} & 97.1 & 10.6 & 20.9 &94.4\\
    \midrule
    \rowcolor[rgb]{0.94, 0.97, 1.0} \multicolumn{5}{l}{\textit{Hierarchical Tokenization}} \\
    \textbf{\ourst-G}& {85.5} & {48.2} & {39.1} &{74.7}\\
    \textbf{\ourst-G} (LLama-2-7b-chat)& {55.1} & \textbf{65.2} & \textbf{46.6} &{49.3}\\
    \textbf{\ourst-GS} & {84.5} & {42.7}& {35.1} & {73.1}\\
    \midrule
    \rowcolor[rgb]{0.94, 0.97, 1.0} \multicolumn{5}{l}{\textit{Ablation variants of \ourst}} \\
    \textbf{\ourst-G w/o \hdata}& {96.6} & 12.5 & 21.6 &96.6\\
    \textbf{\ourst-GS w/o \hdata}& {98.2} & 6.5 & 19.4 &93.3\\
    \bottomrule
\end{tabular}
}
\vspace{-0.2in}
\end{table}

\section{Experimental Evaluation}\label{sec:exp}
We conduct extensive experiments to compare \ours with previous node-centric tokenization across $14$ real-world tasks, including property prediction, molecular description, and chemical reaction prediction. The details and examples regarding the datasets and tasks involved in the experiments are given in Appendix~\ref{appdx:dataset}.
We briefly introduce the setups below and leave the details in Appendix~\ref{appdx:exp}.

\subsection{Experimental settings}
\label{sec:exp_setting}

\textbf{Architecture.} The GNN backbone is a $5$-layer GIN~\citep{gin} with a hidden dimension of $300$. The adapter is a single-layer MLP. We consider base LLMs of  {vicuna-v-1.3-7B}~\citep{vicuna} for all the tasks and {llama-2-7B-chat}~\citep{llama2} for ablation studies.

\textbf{Baselines.} 
Since the focus of this work lies in the tokenization, our main comparison focuses on between \ours and node-centric tokenization. Nevertheless, we also include a series of existing LGLMs based on non-regression LLMs and regression LLMs, to provide an overview of the performance achieved by \ours. We would like to note that there are existing differences in pretraining data and information used between \ours and those baselines. For details, please refer to Table~\ref{tab:compare_other_LGLM} in the Appendix.

For the node-centric based tokenization, we implement the baseline mainly based on InstructMol~\citep{cao2023instructmol} with a VQVAE tokenizer from \molebert~\citep{molebert}.
\ours is implemented based on the same architecture with only the tokenizer replaced. We use the suffix ``-G'' to refer to LGLMs with only 2D graph input and ``-GS'' to refer to LGLMs with both 2D graph and 1D selfies input~\citep{selfies,mol-ins,cao2023instructmol}. We do not include the baselines with ``-GS'' for tasks other than \hbench  as we find that incorporating the 1D input does not always bring improvements in the experiments.

For non-regression-based models, including the pretrained models such as KV-PLM~\citep{kv-plm}, GraphCL~\citep{graphcl} and GraphMVP~\citep{graphmvp}, and molecular foundation models that are trained with tremendous molecule-centric datasets such as MolT5-based methods~\citep{MolT5}, Galactica~\citep{GalacticaAL}, MoMu~\citep{MoMu}, MolFM~\citep{luo2023molfm}, Uni-Mol~\citep{zhou2023unimol}, MolXPT~\citep{molxpt}, GIT-Mol~\citep{liu2024gitmol}, and BioMedGPT~\citep{luo2023biomedgpt}.
We adopt the results from the previous works~\citet{mol-ins,cao2023instructmol} if applicable.

For regression-based LGLMs, we consider LLMs such as ChatGPT~\citep{chatgpt}, Llama~\citep{llama} as well as instruction tuned LLMs such as Alpaca~\citep{alpaca}, Baize~\citep{baize}, ChatGLM~\citep{zeng2023glmb} and Vicuna~\citep{vicuna}. We also consider parameter-efficient finetuned LLMs using the backbone of llama2~\citep{llama2} as done by \molins~\citep{mol-ins}.

\subsection{Motif Hallucination}
\label{sec:hbench_eval}
We begin with a proof-of-concept study with motif hallucination. 
We mainly compare LGLMs with node-centric to that with \ours tokenization with \hbench after stage 1 instruction tuning. For non-regression-based models, we include two state-of-the-art LGLMs  GIMLET~\citep{zhao2023gimlet} and Galactica~\citep{GalacticaAL}.
We do not include the other regression-based models as we found they consistently answered ``Yes'', making a nuanced F1 comparison less informative for them. To avoid the issue of format following, we compare the loss values by feeding the answers of ``Yes'' and ``No'' to the corresponding LLM, calculating the language modeling losses, and taking the one from ``Yes'' and ``No'' with a lower loss as the answer.

\textbf{Reduction of hallucination.} Due to the class imbalance issue in \hbench, we first report comprehensive metrics in Table~\ref{tab:hbench_full}. It can be found that \ours maintains great balance for both positive and negative classes compared to baselines. Especially, in terms of macro F1 scores that are averaged across classes, respectively, \ours demonstrates significant improvements up to $14$\%.

The results of the tokenization-focused comparison are given in Table~\ref{tab:hbench}. Following the practice in LVLMs, we present the F1 scores, accuracies, and the ratio that the model answers ``Yes''~\citep{lvlm_hallu_eval}. Given the imbalance of positive and negative samples, we separately report the F1 scores for different classes. It can be found that the LGLMs with node-centric tokenization consistently answer with ``Yes'' despite the absence of the corresponding functional groups. 
In contrast, \ours significantly reduces the worst class hallucination up to $40$\% in terms of F1 scores, and improves the accuracies up to $30$\%.  The improvements are consistent and significant with both vicunna and llama2 LLM backbones.

\textbf{Ablations with different inputs and LLM backbones.} We also conduct simple ablation studies by additionally incorporating the 1D sequence inputs with SELFIES~\citep{mol-ins,cao2023instructmol}. Contrary to previous results that additionally feeding the 1D sequence always improves the performance of LGLMs, we find that the additional 1D sequence may increase the degree of the hallucination. We suspect that it could be caused by the extremely long sequences of the SELFIES~\citep{selfies} that may distract the attention signals of LLMs.
Nevertheless, \ours still suffers less from the distraction and performs better.

In addition, when without \hdata (or with the \ours architecture), LGLMs will still suffer the hallucination, due to the low quality of the instruction tuning data, demonstrating the necessity of both components of \ours.

\begin{table}
\centering
\small
\caption{\footnotesize
Results of molecular property prediction tasks (regression) on QM9. We report the result in MAE. $\dagger$: few-shot in-context learning (ICL) results from~\citep{mol-ins}. $\Delta{\epsilon}$ refers to the HOMO-LUMO energy gap.
}
\label{tab:property_pred}
\resizebox{0.5\textwidth}{!}{
    \begin{tabular}{lccccc}
        \toprule
        \textsc{Method} &\textsc{HOMO} $\downarrow$ &\textsc{LUMO} $\downarrow$  &$\Delta{\epsilon}$ $\downarrow$ &\textsc{Avg} $\downarrow$\\
        \midrule 
        Alpaca$^\dagger$~\citep{alpaca} & - & - & - & 322.109\\
        Baize$^\dagger$~\citep{baize} & - & - & - & 261.343\\
        LLama2-7B~\citep{llama2} (5-shot ICL) & 0.7367 & 0.8641 & 0.5152 & 0.7510 \\
        Vicuna-13B~\citep{vicuna} (5-shot ICL) & 0.7135 & 3.6807 & 1.5407 & 1.9783 \\
        Mol-Instruction~\citep{mol-ins} & 0.0210 & 0.0210 & 0.0203 & 0.0210 \\
        {InstructMol-G} &  0.0111 & 0.0133 & 0.0147 & 0.0130 \\
        \midrule
        \textbf{\ourst-G} &  \textbf{0.0078} & \textbf{0.0086} & \textbf{0.0095} & \textbf{0.0086} \\
        \bottomrule
    \end{tabular}
}
\vspace{-0.2in}
\end{table}

\subsection{Molecular-Centric Benchmarks}
\textbf{Molecular property prediction} requires LGLMs to answer about particular properties given the molecule.
We use $8$ datasets BACE, BBBP, HIV, SIDER, ClinTox, MUV, and Tox21 from \molnet, and CYP450 from GIMLET~\citep{zhao2023gimlet} to evaluate the classification performance with ROC-AUC. 
We also adopt the regression-based property prediction datasets from~\citep{mol-ins}, where we evaluate several quantum chemistry measures such as HUMO, LUMO, and HUMO-LUMO gap~\citep{Ramakrishnan2014QuantumCS} via Mean Absolute Error (MAE). 

The results of molecular property prediction are given in Table~\ref{tab:property_pred} and Table~\ref{tab:molnet_full} for regression and classification, respectively.
We can find that \ours always significantly boosts the performance in both types of tasks. 
Remarkably, in CYP450~\citep{zhao2023gimlet}, \ours significantly outperforms the state-of-the-art model, demonstrating the advances of LGLM with hierarchical graph tokenization.
Interestingly, Llama-2~\citep{llama2} can match the state-of-the-art performance in HIV in a few-shot setting, while performing significantly worse in other datasets, for which we suspect some data contamination might exist.

\begin{table*}[ht]
\centering
\small
\caption{\footnotesize
ROC-AUC Results of molecular property prediction tasks (classification) on MoleculeNet~\citep{molnet}. Evaluation on InstructMol and \ours adopt the likelihood of the tokens of ``Yes'' and ``No''. Most of the instruction tuning datasets are from GIMLET~\citep{zhao2023gimlet}. SIDER and ClinTox are converted following the MoleculeNet task description.
}
\resizebox{\textwidth}{!}{
\begin{tabular}{lcccccccc}
    \toprule
    \textsc{Method} &BACE $\uparrow$  &BBBP $\uparrow$ &HIV $\uparrow$ &SIDER $\uparrow$  &ClinTox &MUV $\uparrow$ &Tox21 $\uparrow$&CYP450 $\uparrow$  \\
    \textsc{\# Molecules} &1,513 &2,039 &41,127 &1,427&1,478&93,087&7,831&16,896\\
    \textsc{\# Tasks} &1 &1 &1 &27&2&17&12&5\\
    \midrule
    KV-PLM~\citep{kv-plm} &78.5 &70.5 &71.8 &59.8&84.3&61.7&49.2&59.2\\
    GraphCL~\citep{graphcl} &75.3 &69.7 &78.5&60.5&76.0&69.8&73.9&- \\
    GraphMVP-C~\citep{graphmvp} &81.2 &72.4 &77.0&60.6&84.5&74.4&77.1&- \\
    MoleculeSTM-G~\citep{liu2023moleculestm} &80.8&70.0&76.9&61.0&92.5&73.4&76.9&- \\
    MoMu~\citep{MoMu} &76.7 &70.5 & 75.9 &60.5&79.9&60.5&57.8&58.0\\
    MolFM~\citep{luo2023molfm} &83.9 &\textbf{72.9} &78.8&64.2&79.7&76.0&77.2&- \\
    Uni-Mol~\citep{zhou2023unimol} &\textbf{85.7} &\textbf{72.9 }&\textbf{80.8} &65.9&91.9&82.1&78.1&-\\
    Galactica-1.3B~\citep{GalacticaAL} & 57.6 & 60.4 & 72.4&54.0&58.9&57.2&60.6&46.9 \\
    Galactica-6.7B~\citep{GalacticaAL} & 58.4 & 53.5 & 72.2&55.9&78.4&-&63.9&- \\
    Galactica-30B~\citep{GalacticaAL}  & 72.7 & 59.6 & \textbf{75.9} &61.3&82.2&-&68.5&-\\
    Galactica-120B~\citep{GalacticaAL} & 61.7 & 66.1 & 74.5 &63.2&82.6&-&68.9&-\\
    GIMLET~\citep{zhao2023gimlet} & 69.6 & 59.4 & 66.2 &-&-&64.4&61.2&71.3\\
    \midrule 
    
    LLama-2-7b-chat (4-shot)~\citep{llama2} & 76.9 & 54.2 & 67.8 &- &- &46.9&62.0 &57.6\\
    LLama-2-13b-chat (4-shot)~\citep{llama2} & 74.7 & 52.8 & \textbf{72.4} &- &- &47.9&57.5&55.6 \\
    {InstructMol-G} & 64.3 & 48.7 & 50.2 & 51.0 & 50.0 & 50.0& 59.0 & 59.1\\
    \textbf{\ourst-G}& \textbf{77.1} & \textbf{61.8} & {63.3} & \textbf{58.8} & \textbf{55.3} & \textbf{51.1}& \textbf{67.4}&\textbf{80.5} \\
    \bottomrule
\end{tabular}}
\label{tab:molnet_full}
\end{table*}

\begin{table*}[t]
\centering
\vspace{-0.15in}
\small
\caption{\footnotesize
Results of molecular description generation task on the test split of ChEBI-20.
}
\label{tab:molcap}
\resizebox{\textwidth}{!}{
\begin{tabular}{lcccccc}
\toprule
\textsc{Model}
&\textsc{BLEU-2}$\uparrow$  & \textsc{BLEU-4}$\uparrow$  & \textsc{ROUGE-1}$\uparrow$  & \textsc{ROUGE-2}$\uparrow$  & \textsc{ROUGE-L}$\uparrow$ & \textsc{METEOR}$\uparrow$ \\

\midrule[1.1pt]
MoT5-base~\citep{MolT5}        & 0.540 &0.457 &0.634 &0.485 &0.568 &0.569 \\
MoMu (MolT5-base)~\citep{MoMu} & 0.549 &0.462 &  -    &  -    &  -    &0.576 \\
MolFM (MolT5-base)~\citep{luo2023molfm}&0.585 &0.498 &0.653 &0.508 &0.594 &0.607 \\
MolXPT~\citep{molxpt}          &0.594  &0.505 &0.660 &0.511 &0.597 &0.626 \\
GIT-Mol-graph~\citep{liu2024gitmol}  & 0.290 &0.210 &0.540 &0.445 &0.512 & 0.491 \\
GIT-Mol-SMILES~\citep{liu2024gitmol} & 0.264 &0.176 &0.477 &0.374 &0.451 &0.430  \\
GIT-Mol-(graph+SMILES)~\citep{liu2024gitmol} & 0.352 & 0.263 &0.575 &0.485 &0.560 &0.430 \\
Text+Chem T5-augm-base~\citep{Text+ChemT5} &\textbf{0.625} &\textbf{0.542} &\textbf{0.682} &\textbf{0.543} &\textbf{0.622} &\textbf{0.648} \\
GPT-3.5-turbo (10-shot MolReGPT)~\citep{molregpt} &0.565 &0.482 &0.623 &0.450 &0.543 &0.585 \\
GPT-4-0314 (10-shot MolReGPT)~\citep{molregpt} &0.607 &0.525 &0.634 &0.476 &0.562 &0.610 \\
\midrule
GPT-3.5-turbo (zero-shot)~\citep{molregpt} & 0.103 &0.050 &0.261 &0.088 &0.204 &0.161 \\
BioMedGPT-10B~\citep{luo2023biomedgpt}          &0.234 &0.141  &0.386  &0.206  &0.332  &0.308  \\
Mol-Instruction~\citep{mol-ins} &0.249 &0.171  &0.331  &0.203  &0.289  &0.271  \\
{InstructMol-G}         &0.481	&0.381	&0.554	&0.379	&0.488	&0.503	\\
\midrule
\textbf{\ourst-G}         &\textbf{0.504}	&\textbf{0.405}	&\textbf{0.570}	&\textbf{0.397}	&\textbf{0.502}	&\textbf{0.524}	\\
\bottomrule
\end{tabular}
}
\end{table*}

\textbf{Molecular description} requires the LGLMs to generate a caption of the molecule.
We adopt the widely used benchmark \chebi~\citep{edwards2021text2mol} which evaluates the linguistic distances of the generated molecule captions of molecular characteristics such as structure, properties, biological activities etc.. 
We report the metrics of BLEU~\citep{papineni-etal-2002-bleu}, ROUGE~\citep{lin-2004-rouge} and Meteor~\citep{banerjee-lavie-2005-meteor}. 
The LGLMs are trained using the \chebi train split, selected according to the best training loss, and evaluated using the test split.

As shown in Table~\ref{tab:molcap}, \ours consistently brings significant improvements over LGLMs with node-centric tokenization. Nevertheless, compared to the molecular foundation models such as MoT5~\citep{MolT5} pretrained on a significant amount of molecule-text related corpus, there remains a gap for regression-based LGLMs even with \ours.
The gap calls for future investigations on how to incorporate \ours into the pretraining of the LGLMs properly.

\begin{table*}[t]
\centering
\small
\caption{\footnotesize
Results of chemical reaction tasks. These tasks encompass reagent prediction, forward reaction prediction, and retrosynthesis. $\dagger$: few-shot ICL results from \cite{mol-ins}. $*$: use task-specific instruction data to finetune.
}
\label{tab:chemical_reaction}
\resizebox{\textwidth}{!}{
\begin{tabular}{lccccccc}
\toprule
\textsc{Model}
&\textsc{Exact}$\uparrow$  & \textsc{BLEU}$\uparrow$  & \textsc{Levenshtein}$\downarrow$  & \textsc{RDK FTS}$\uparrow$  & \textsc{MACCS FTS}$\uparrow$ & \textsc{Morgan FTS}$\uparrow$ & \textsc{Validity}$\uparrow$ \\

\midrule[1.1pt]
\rowcolor[rgb]{0.94, 0.97, 1.0}
\multicolumn{8}{l}{\textit{Reagent Prediction}} \\
Alpaca$^\dagger$~\citep{alpaca} & 0.000 & 0.026 & 29.037 & 0.029 & 0.016 & 0.001 & 0.186 \\
Baize$^\dagger$~\citep{baize} & 0.000 & 0.051 & 30.628 & 0.022 & 0.018 & 0.004 & 0.099 \\
ChatGLM$^\dagger$~\citep{zeng2023glmb} & 0.000 & 0.019 & 29.169 & 0.017 & 0.006 & 0.002 & 0.074 \\
Llama$^\dagger$~\citep{llama} & 0.000 & 0.003 & 28.040 & 0.037 & 0.001 & 0.001 & 0.001 \\
Vicuna$^\dagger$~\citep{vicuna} & 0.000 & 0.010 & 27.948 & 0.038 & 0.002 & 0.001 & 0.007 \\
Mol-Instruction~\citep{mol-ins} & 0.044 & 0.224 & \textbf{23.167} & 0.237 & \textbf{0.364 }& 0.213 & 1.000 \\
Llama-7b$^*$~\citep{llama}(LoRA) &0.000	&0.283	&53.510	&0.136	&0.294	&0.106	&1.000 \\
{InstructMol-G}       & 0.031	&0.429	&31.447	&0.389	&0.249	&0.220	&1.000\\
\midrule
\textbf{\ourst-G}       & 0.050	&\textbf{0.462}	&28.970	&\textbf{0.441}	&0.314	&\textbf{0.275}	&1.000\\
\midrule[1.1pt]
\rowcolor[rgb]{0.94, 0.97, 1.0}
\multicolumn{8}{l}{\textit{Forward Reaction Prediction}} \\
Alpaca$^\dagger$~\citep{alpaca} & 0.000 & 0.065 & 41.989 & 0.004 & 0.024 & 0.008 & 0.138 \\
Baize$^\dagger$~\citep{baize} & 0.000 & 0.044 & 41.500 & 0.004 & 0.025 & 0.009 & 0.097 \\
ChatGLM$^\dagger$~\citep{zeng2023glmb} & 0.000 & 0.183 & 40.008 & 0.050 & 0.100 & 0.044 & 0.108 \\
Llama$^\dagger$~\citep{llama} & 0.000 & 0.020 & 42.002 & 0.001 & 0.002 & 0.001 & 0.039 \\
Vicuna$^\dagger$~\citep{vicuna} & 0.000 & 0.057 & 41.690 & 0.007 & 0.016 & 0.006 & 0.059 \\
Mol-Instruction~\citep{mol-ins} & 0.045 & 0.654 & 27.262 & 0.313 & 0.509 & 0.262 & 1.000 \\
Llama-7b$^*$~\citep{llama}(LoRA) &0.012	&0.804	&29.947	&0.499	&\textbf{0.649}	&\textbf{0.407}	&1.000 \\
{InstructMol-G}        & 0.031 &0.853 &24.790	&0.512	&0.362	&0.303	&0.993 \\
\midrule
\textbf{\ourst-G}        & 0.037 &\textbf{0.869} &\textbf{23.759}	&\textbf{0.590}	&0.394	&0.340	&0.993 \\
\midrule[1.1pt]
\rowcolor[rgb]{0.94, 0.97, 1.0}
\multicolumn{8}{l}{\textit{Retrosynthesis}} \\
Alpaca$^\dagger$~\citep{alpaca} & 0.000 & 0.063 & 46.915 & 0.005 & 0.023 & 0.007 & 0.160 \\
Baize$^\dagger$~\citep{baize} & 0.000 & 0.095 & 44.714 & 0.025 & 0.050 & 0.023 & 0.112 \\
ChatGLM$^\dagger$~\citep{zeng2023glmb} & 0.000 & 0.117 & 48.365 & 0.056 & 0.075 & 0.043 & 0.046 \\
Llama$^\dagger$~\citep{llama} & 0.000 & 0.036 & 46.844 & 0.018 & 0.029 & 0.017 & 0.010 \\
Vicuna$^\dagger$~\citep{vicuna} & 0.000 & 0.057 & 46.877 & 0.025 & 0.030 & 0.021 & 0.017 \\
Mol-Instruction~\citep{mol-ins} & 0.009 & 0.705 & 31.227 & 0.283 & \textbf{0.487} & 0.230 & 1.000 \\
Llama-7b$^*$~\citep{llama}(LoRA) &0.000	&0.283	&53.510	&0.136	&0.294	&0.106	&1.000 \\
{InstructMol-G}       & 0.001	&0.835	&31.359	&0.447	&0.277	&0.241	&0.996\\
\midrule
\textbf{\ourst-G}       & 0.008	&\textbf{0.863}	&\textbf{28.912}	&\textbf{0.564}	&0.340	&\textbf{0.309}	&1.000\\
\bottomrule
\end{tabular}
}
\vspace{-0.15in}
\end{table*}

\textbf{Chemical reaction prediction} requires the LGLMs to predict the results of the chemical reaction analysis, which are crucial for AI-aided drug discovery~\citep{mol-ins}. 
Reagent prediction aims to predict the suitable reagents for a particular chemical reaction. Forward reaction prediction aims to predict the products of a chemical reaction, given the reactants and the reagents.
Retrosynthesis prediction aims to predict the suitable reactants given a target product.
The inputs and outputs for chemical reaction related tasks adopt the SELFIES~\citep{selfies} as recommended by~\citep{mol-ins}.
We report both linguistic distance metrics such as BLEU~\citep{papineni-etal-2002-bleu} and Levenshtein~\citep{levenshtein}, and molecular similarity measures such as similarity of the molecular fingerprints~\citep{rdkit}.

As shown in Table~\ref{tab:chemical_reaction}, across all tasks in chemical reaction prediction, LGLMs with \ours consistently and significantly improve the performances compared to the node-centric tokenization. Meanwhile, LGLMs with \ours achieve state-of-the-art results in several tasks and metrics, compared to other regression-based LGLMs that even incorporate a stronger LLM backbone such as Mol-Instruction, and additional information of SELFIES.

\begin{figure*}[h!]
    \centering
    \subfigure[Different training settings]{
        \includegraphics[width=0.3\textwidth]{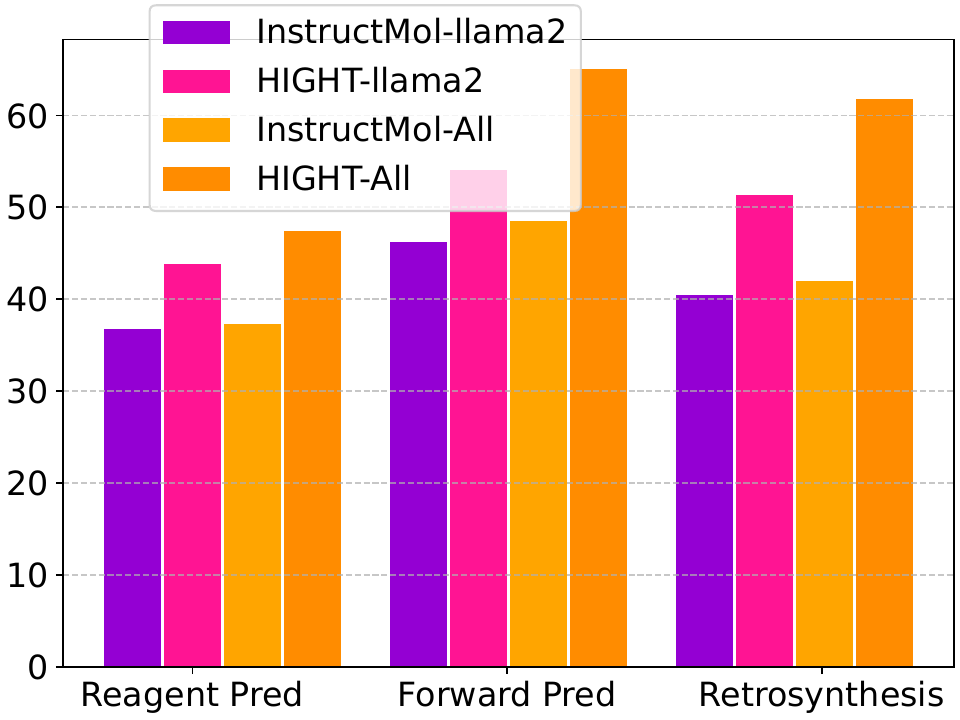}
        \label{fig:llama2b}
    }
    \subfigure[Zero-shot transfer]{
        \includegraphics[width=0.3\textwidth]{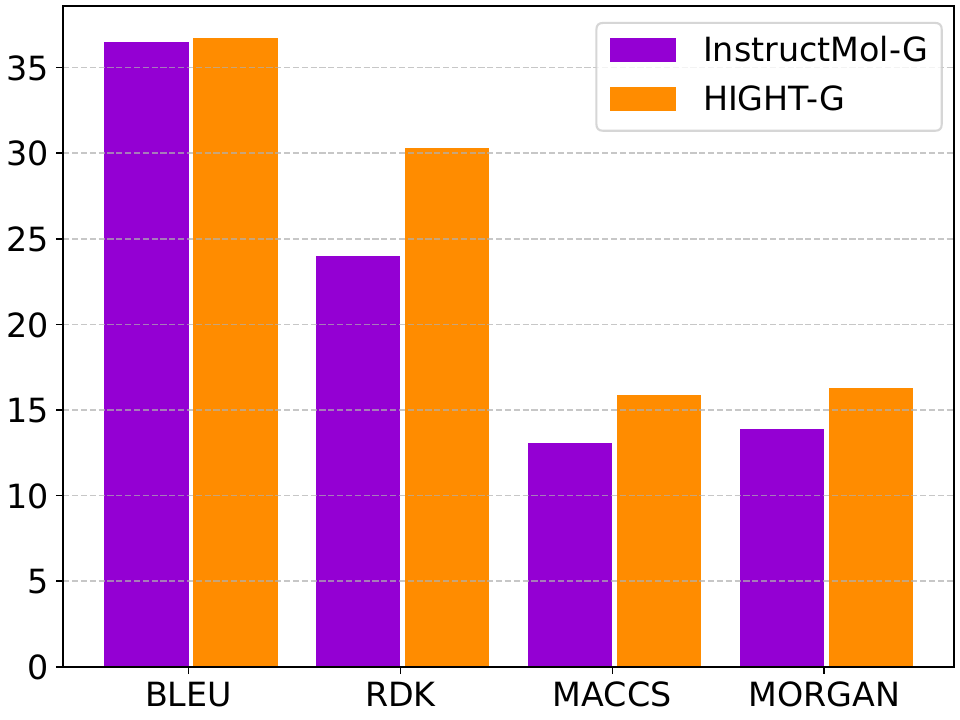}
        \label{fig:transfer}
    }
    \subfigure[Ablation variants of \ours]{
        \includegraphics[width=0.3\textwidth]{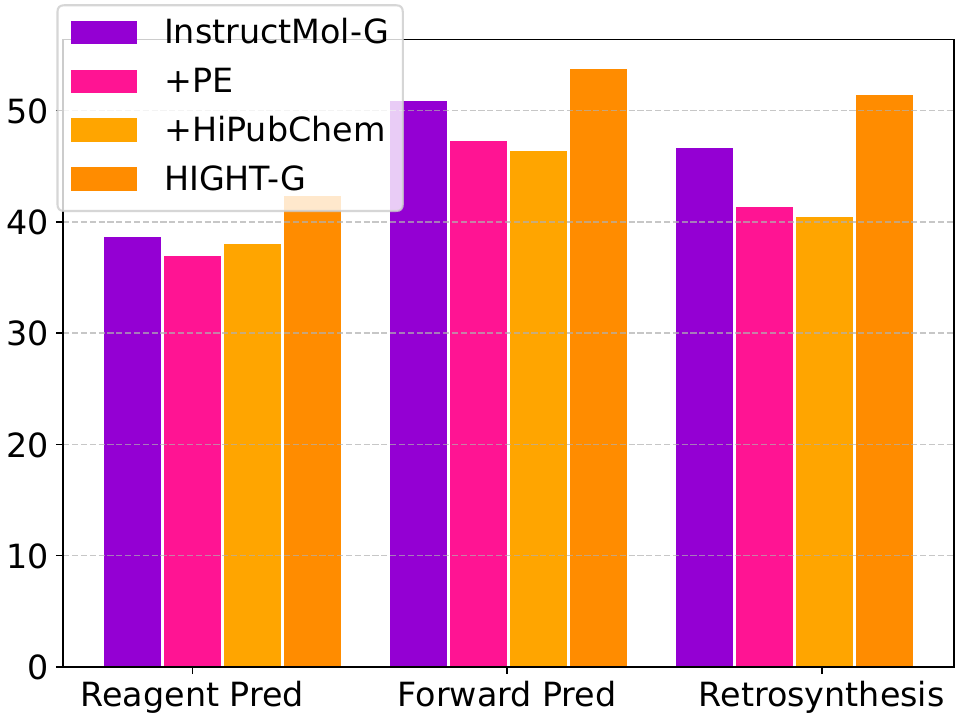}
        \label{fig:ablation}
    }
    \vspace{-0.1in}
    \caption{Ablation studies.}
    \label{fig:ablation_exp_all}
    \vspace{-0.15in}
\end{figure*}

\subsection{Empirical Analysis}
\textbf{Generalist capabilities.} We follow the previous practice in training and evaluating generalist models~\citep{liu2023llava} and consider the two settings:
a) As shown in Fig.~\ref{fig:llama2b}, we first train the model with all chemical reaction prediction data by $3$ epochs to elicit the format following and the knowledge adaption capabilities of the LGLMs after stage 1. The models are named with ``-All'';
b) As shown in Fig.~\ref{fig:transfer}, we train the model with retrosynthesis task data and evaluate the zero-shot transfer performance on forward reaction prediction.
Under both settings, we can find that \ours boosts the generalist capabilities significantly.

\textbf{Computation overhead.} In Appendix~\ref{appdx:computation}, we report the computation overhead of pretraining and inference as well as tunable parameters of \ours and InstructMol. It can be found that, although \ours requires longer training time and relatively higher tunable parameters, the absolute values are not high. Moreover, during inference, as LLM latency consumes most of the computation, \ours can even reduce the inference latency by generating more concise answers.

\textbf{Ablation studies.} To better understand the effectiveness of distinct components in \ours,
we conduct ablation studies that train InstructMol~\citep{cao2023instructmol} with the laplacian positional encodings or with \hdata, as given in Fig.~\ref{fig:ablation}. We can find that, merely incorporating positional encoding or hierarchical instruction tuning is not sufficient to achieve the same performance as \ours. On the contrary, without a proper architecture design as \ours, LGLMs with previous node-centric tokenization with \hdata will confuse LLMs and even lead to degenerated downstream task performances. In addition, we also compare LGLMs with llama2 backbone. As shown in Fig.~\ref{fig:llama2b}, \ours still significantly boosts the performance. More ablation studies are provided in Appendix~\ref{appdx:ablation}.

\section{Conclusions}
This paper presents \ours, a novel hierarchical graph tokenization technique. By incorporating the hierarchical graph information, \ours improves the molecule-language alignment performance, reducing hallucinations and boosting accuracy in molecular tasks.
Nevertheless, the current focus on molecular graphs requires further verification for wider applicability to other forms of graph data, such as those originating from social networks. 
Despite the limitation, \ours represents a significant step forward in advancing graph comprehension capability of LLMs, and highlighting paths for future research in this direction.

Meanwhile, incorporating 3D information into the graph-language alignment is also a promising future direction, especially for broader scientific tasks such as single-cell modeling and understanding. For example, built upon \ours, one could design a new 3D tokenizer to accommodate 3D properties of motifs, scale up 3D data to include amino acids in proteins and certain recurrent structures in RNA sequences, incorporate 3D  positional encoding, and curate instruction tuning data with 3D descriptive captions.

\clearpage

\section*{Acknowledgments}
We thank the reviewers for their valuable comments. JC was supported by RGC Young Collaborative Research Grant No. C2005-24Y.

\section*{Impact Statement}
This paper mainly focuses on how to best represent graph information for LLMs to better understand the graphs. We demonstrate the effectiveness of our method on molecule-centric tasks, which could facilitate the broader use of LLMs for tasks like AI-aided drug discovery and human-machine interactions in biomedicine.
Besides, this paper does not raise any ethical concerns.
This study does not involve
any human subjects, practices, to data set releases, potentially harmful insights, methodologies and
applications, potential conflicts of interest and sponsorship, discrimination/bias/fairness concerns,
privacy and security issues, legal compliance, and research integrity issues.

\bibliography{references/llm,references/graphllm,references/mllm,references/graphood,references/ood,references/references,references/xgnn}
\bibliographystyle{icml2025}

\clearpage
\onecolumn
\appendix

\begin{center}
    \LARGE \bf {Appendix of \ourst}
\end{center}
\etocdepthtag.toc{mtappendix}
\etocsettagdepth{mtchapter}{none}
\etocsettagdepth{mtappendix}{subsection}
\tableofcontents

\clearpage

\section{More Future Works}
Built upon \ours, there are several promising future directions.
For example, one could extend this study to more types of graphs, such as social networks and knowledge graphs, by exploring the crucial substructures therein:
\begin{itemize}
    \item Indeed, motifs generically exist in other types of graphs and are crucial for a variety of tasks~\citep{10.1145/3433652}. For example, cliques can define boundaries between groups of people in social networks~\citep{DOREIAN1994267}. The idea of HIGHT could be seamlessly applied to other graphs where we have some prior knowledge about critical motifs.
    \item Meanwhile, when we do not have prior knowledge about the motifs, the GNNs intrinsically model the hierarchical nature of graphs in different orders~\citep{diffpool} and thus can be integrated into LGLMs to learn the hierarchical graph information. A similar idea has been verified successful in graph transformers~\citep{zhang2022hierarchical}.
    \item Furthermore, one could also adopt interpretable GNNs to identify the critical subgraphs for the task~\citep{gib,gsat,chen2024gmt} that capture the underlying causal information about the underlying tasks~\citep{ciga,chen2023gala,yao2024empowering,causalcoat2024,Yao2025LearningGI,Xu2025BrainOODOG}.
    It is also interesting to further investigate the hallucinations caused by the spurious correlations during the alignment~\citep{counteranimal}.
\end{itemize}

\section{Comparison between other LGLMs}

\begin{table}[H]
\caption{Comparison between other LGLMs in terms of the backbone, instruction tuning, downstream usage for Molecular Property prediction, and capable tasks. It can be found that HIGHT is capable of various tasks, given limited pre-training data and information. Note that compared to the instruction tuning data for other LGLMs, such as KV-PLM~\citep{kv-plm}, which consists of papers with detailed information about molecules, the text descriptions in HIGHT contain relatively simple sentences.}
\label{tab:compare_other_LGLM}
\resizebox{\textwidth}{!}{
\begin{tabular}{llllllr}
\toprule
Model                                       & Backbone    & Information                      & Instruction Tuning   Data               & Downstream  & \# Tasks \\\midrule
HIGHT/InstructMol~\citep{cao2023instructmol}        & GNN+Llama   & 2D Graph+Text                    & HiPubChem-295k                      & LoRA                  & 7                                   \\
GraphCL~\citep{graphcl}                  & GNN         & 2D Graph                         &Downstream training data        & Finetuning            & 1                                   \\
GraphMVP~\citep{graphmvp}                 & GNN         & 2D+3D graph                      & Geom-50K                            & Finetuning            & 1                                   \\
MoleculeSTM~\citep{molebert}              & GNN+SciBERT & 2D Graph+3D Graph+Text           & PubChemSTM-280K                     & Finetuning            & 6                                   \\
KV-PLM~\citep{kv-plm}                  & BERT        & 1D SMILES+Text                   & S2orc-300K academic papers          & Finetuning            & 5                                   \\
MolT5~\citep{MolT5}                & T5          & 1D SMILES+Text                   & C4+ZINC(100M)                       & Finetuning            & 2                                   \\
Text+Chem T5~\citep{Text+ChemT5} & T5          & 1D SMILES+Text                   & Multi task-33.5M                    & Finetuning            & 5                                   \\
MoMu~\citep{MoMu}                      & GNN+BERT    & 2D Graph+Text                    & Graph-Docuemnt Pair-15.6K           & Finetuning            & 4                                   \\
MolFM~\citep{luo2023molfm}                    & GNN+BERT    & Knowledge Graph+2D+3D Graph+Text & KG-15K+S2ORC-37M                    & Finetuning            & 4                                   \\
Uni-Mol~\citep{zhou2023unimol}                 & Transformer & 3D Graph                         & molecule-209M+protein-3.2M          & Finetuning            & 4                                   \\
Galactica~\citep{GalacticaAL}             & GPT         & 1D SMILES/Text                   & Documents-59M+chemicla prompts-2.5M & zero-shot                                     & 12                                  \\
MolXPT~\citep{molxpt}                  & GPT2        & 1D SMILES+Text                   & Mixed text-68M                      & Finetuning            & 2                                   \\
GIMLET~\citep{zhao2023gimlet}                   & GNN+T5      & 2D Graph+Text                    & ChemBL-730K                         & zero-shot                                     & 2          \\\bottomrule                       
\end{tabular}}
\end{table}

\section{Details of Instruction Tuning Datasets}
\label{appdx:dataset}
We provide a summary of the datasets for instruction tuning and evaluation in this paper as in Table~\ref{tab:dataset_summary}.
Meanwhile, we also list the data sources and the corresponding licenses of the sources for each task and dataset. Then, we will elaborate more on the details of the datasets in the following subsections. 

\bgroup
\def\arraystretch{1.2}
\begin{table*}[!ht]
    \centering
    \caption{
    \small
    Summary of datasets involved in our paper.
    }
    \resizebox{\textwidth}{!}{
    \begin{tabular}{p{4cm}|p{1cm}p{1cm}p{8cm}}
    \toprule
    \textbf{Datasets} &\textbf{Train} & \textbf{Test}&\textbf{Content}  \\
    \midrule
    PubChem&295,228 &N/A & Molecules and the associated descriptions from PubChem. \\
    \hdata &295,228 &N/A & Molecules and the associated descriptions from PubChem and about functional groups in the molecule. \\
    MoleculeNet-HIV&32,901 & 4,113 & Question answering about the ability of the molecule to inhibit HIV replication.\\
    MoleculeNet-BACE&1,210 & 152 & Question answering about the ability of the molecule to bind to the BACE1 protein \\
    MoleculeNet-BBBP&1,631 & 204 & Question answering about the ability of the molecule to diffuse across the brain blood barrier. \\
    MoleculeNet-SIDER&1,141 & 143 & Question answering about the ability of the side effects. \\
    MoleculeNet-ClinTox&1,188 & 148 & Question answering about the toxicology. \\
    MoleculeNet-MUV&74,469 & 9,309 & Question answering about PubChem bioAssay \\
    MoleculeNet-Tox21&6,877 & 860 & Question answering about Toxicology in the 21st century \\
    CYP45-&13,516 & 1,690 & Question answering about CYP PubChem BioAssay CYP 1A2, 2C9, 2C19, 2D6, 3A4 inhibition. \\
    Property Prediction (Regression)&360,113& 1,987 & Question answering about the quantum mechanics properties of the molecule. \\
    Forward Reaction Prediction&124,384 & 1,000 & Question answering about the products of a chemical reaction, given specific reactants and reagents. \\
    Reagent Prediction&124,384 & 1,000 & Question answering about suitable catalysts, solvents, or ancillary substances required for a specific chemical reaction. \\
    Retrosynthesis Prediction&128,684 & 1,000 & Question answering about the reactants and reagents of a chemical reaction, given specific products. \\
    ChEBI-20 &26,407 & 3,300 & Molecules and the associated Chemical Entities of Biological Interest (ChEBI)~\citep{chebi} annotations. \\
    \hbench &N/A & 23,924 & Question answering about existing functional groups in the molecule. \\
    \bottomrule
    \end{tabular}}
    \label{tab:dataset_summary}
\end{table*}
\egroup

\begin{table*}[!ht]
    \centering
    \caption{
    \small
    Summary of data resources and licenses of datasets involved in our paper.
    }
    \scalebox{0.69}{
    \begin{tabular}{p{4cm}|p{4cm}p{5cm}p{5cm}}
    \toprule
    \textbf{Tasks/Datasets} &\textbf{Data Sources} & \textbf{License URL} & \textbf{License Note} \\
    \midrule
    PubChem, \hdata &PubChem & \url{https://www.nlm.nih.gov/web_policies.html} & Works produced by the U.S. government are not subject to copyright protection in the United States. Any such works found on National Library of Medicine (NLM) Web sites may be freely used or reproduced without permission in the U.S. \\
    Reaction Prediction&USPTO & \url{https://www.uspto.gov/learning-and-resources/open-data-and-mobility} & It can be freely used, reused, and redistributed by anyone. \\
    Property Prediction&MoleculeNet & \url{https://opensource.org/license/mit/} & Permission is hereby granted, free of charge, to any person obtaining a copy of this software and associated documentation files (the “Software”), to deal in the Software without restriction, including without limitation the rights to use, copy, modify, merge, publish, distribute, sublicense, and/or sell copies of the Software, and to permit persons to whom the Software is furnished to do so. \\
    Property Prediction&CYP450 & \url{https://www.nlm.nih.gov/web_policies.html} & The data is from~\citet{zhao2023gimlet} that curates PubChem BioAssay CYP 1A2, 2C9, 2C19, 2D6, 3A4 inhibition. Thus it shares the same license as PubChem. \\
   Molecular Description, \hbench & ChEBI & \url{https://creativecommons.org/licenses/by/4.0/} & You are free to: Share — copy and redistribute the material in any medium or format. Adapt — remix, transform, and build upon the material for any purpose, even commercially. \\
    \bottomrule
    \end{tabular}
    }
    \label{tab:licence}
\end{table*}

\begin{table}[ht]
\caption{
    \small
    {Summary of inputs and outputs of the tasks in experiments.}
    }
\resizebox{\textwidth}{!}{
\begin{tabular}{lll}\toprule
                                               & input                                                     & output            \\\midrule
motif hallucination                            & molecule and question about the existence of a motif      & yes or no         \\
molecular property prediction (classification) & molecule and question about the existence of the property & yes or no         \\
molecular property prediction (regression)     & molecule and question about the value of the property     & property value    \\
molecular caption                              & molecule and question asking for the molecular caption    & molecular caption \\
chemical reaction prediction                   & molecules and question about the reaction                 & molecular results\\\bottomrule
\end{tabular}}
\end{table}

\subsection{Details of the PubChem Dataset}
\label{appdx:pubchem}
PubChem\footnote{\url{https://pubchem.ncbi.nlm.nih.gov}} is one of the largest public molecule database~\citep{pubchem}, and has been widely adopted by the alignment training of LGLMs~\citep{liu2023molca,liu2023moleculestm,cao2023instructmol}. Our construction of PubChem predominantly follows~\citet{liu2023moleculestm}. We will briefly describe the main steps and interested readers may refer the details to~\citep{liu2023moleculestm}:
\begin{itemize}
    \item We curate the data from PubChem using the official API and set the data cutoff date as 12 Jan. 2024. It downloads both the molecular structure (e.g., SMILES, 2D molecular graphs) in SDF format, and the text descriptions.
    \item Then, we will filter out molecules that do not have descriptions or can not match via the PubChem ID. In the descriptions, the molecule names are replaced with ``This molecule'', in order to facilitate LLMs to understand the instructions.
\end{itemize}
Finally, the curation generates $295$k molecule-text pairs that we term as PubChem-$295$k. PubChem-$295$k will be mainly used for the stage 1 alignment training.

\begin{table*}[!ht]
    \centering
    \caption{
    \small
    Examples of PubChem and \hdata datasets.
    }
    \scalebox{0.69}{
    \begin{tabular}{p{10cm}p{10cm}}
    \toprule
    \textbf{PubChem} &\textbf{HiPubChem} \\
    \midrule  
    \textit{SMILES: CC(=O)OC(CC(=O)[O-])C[N+](C)(C)C}&\\
    This molecule is an O-acylcarnitine having acetyl as the acyl substituent. It has a role as a human metabolite. It is functionally related to an acetic acid. It is a conjugate base of an O-acetylcarnitinium. 
    & This molecule has 1 carboxylic acids functional group. This molecule has no methyl amide, or amide, or nitro or thiols groups. This molecule is an O-acylcarnitine having acetyl as the acyl substituent. It has a role as a human metabolite. It is functionally related to an acetic acid. It is a conjugate base of an O-acetylcarnitinium. \\
    \textit{SMILES: CCN(CC)CCOC(=O)C(Cc1cccc2ccccc12)CC1CCCO1}&\\
    This molecule is a member of naphthalenes.
    & This molecule has 0 functional groups. This molecule is a member of naphthalenes. \\
    \textit{SMILES: Cc1c2[nH]c(c1CCC(=O)O)Cc1[nH]c(c(CCC(=O)O)c1C)Cc1[nH]c(c(CCC(=O)O)c1C)Cc1[nH]c(c(C)c1CCC(=O)O)C2}&\\
    This molecule is a coproporphyrinogen. It has a role as an Escherichia coli metabolite and a mouse metabolite. It is a conjugate acid of a coproporphyrinogen III(4-).
    & This molecule has 1 carboxylic acids functional groups. This molecule has no methyl amide, or diazo, or cyano or thiols groups. This molecule is a coproporphyrinogen. It has a role as an Escherichia coli metabolite and a mouse metabolite. It is a conjugate acid of a coproporphyrinogen III(4-). \\
    \bottomrule
    \end{tabular}
    }
    \label{tab:example_pubchem}
\end{table*}

\subsection{Details of \hdata Dataset}
\label{appdx:hdata}

\hdata augments the molecular instruction tuning dataset with captions of the functional groups. 
We consider both the positive and negative appearances of motifs when augmenting the instructions. For the positive case, we directly append the caption of all functional groups detected with RDKit:
\begin{myquotation}\centering
    \texttt{This molecule has <\#> of <functional group name> groups.}
\end{myquotation}
For the negative case, we randomly sample $k_\text{neg}$ that do not appear in the molecule:
\begin{myquotation}\centering
    \texttt{This molecule has no <functional group name> groups.}
\end{myquotation}
Despite the simple augmentation strategy, we find that \hdata significantly reduces the hallucination issue, and improves the molecule-language alignment performance.

For comparison, we provide examples of PubChem and \hdata in Table~\ref{tab:example_pubchem}.

\subsection{Details of Property Prediction Dataset}
\label{appdx:property_pred}
The task of molecular property prediction mainly aims to predict certain biochemical or physical properties of molecules. Usually, these properties have a close relation with the molecular substructures (i.e., functional groups)~\citep{art_drug}. In this work, we consider the scenarios of both binary classification based and the regression based molecular property prediction, and the datasets are mainly derived from MoleculeNet~\citep{molnet}.

For the classification, we consider three subtasks, HIV, BACE, and BBBP. The HIV subtask mainly evaluates whether the molecule is able to impede the replication of the HIV virus. The BACE subtask mainly evaluates the binding capability of a molecule to the BACE1 protein. The BBBP subtask mainly evaluates the capability of a molecule to passively diffuse across the human brain blood barrier. For task-specific instruction tuning, we convert those classification based datasets into instructions. Examples are given in Table~\ref{tab:example_proprety_bin}.
\bgroup
\def\arraystretch{1.2}
\begin{table*}[!ht]
    \centering
    \caption{
    \small
    Examples of the property prediction (classification) datasets.
    }
    \scalebox{0.7}{
    \begin{tabular}{p{2cm}p{14cm}p{2cm}}
    \toprule
    \textbf{Dataset}&\textbf{Question} &\textbf{Answer} \\
    \midrule
    HIV&\textit{SMILES: N=C1OC2(c3ccccc3)C3=C(OC(=NC)N2C)C(=O)OC3(c2ccccc2)N1C}&\\
    &Please help me evaluate whether the given molecule can impede the replication of the HIV virus. 
    & No \\
    BACE&\textit{SMILES: CN(C(=O)CCc1cc2ccccc2nc1N)C1CCCCC1}&\\
    &Can the given molecule bind to the BACE1 protein? 
    & Yes \\
    BBBP&\textit{SMILES: Cc1c[nH+][o+]c(C([NH])CC(C)C(C)(C)N(C(C)(C)C)C(C)(N)N)c1[O-]}&\\
    &Can the given molecule passively diffuse across the brain blood barrier? 
    & Yes \\
    \bottomrule
    \end{tabular}
    }
    \label{tab:example_proprety_bin}
\end{table*}
\egroup

\bgroup
\def\arraystretch{1.2}
\begin{table*}[!ht]
    \centering
    \caption{
    \small
    Examples of the property prediction (regression) datasets.
    }
    \scalebox{0.8}{
    \begin{tabular}{p{14cm}c}
    \toprule
    \textbf{Question} &\textbf{Answer} \\
    \midrule
    \textit{SELFIES: [O][=C][O][C][C][C][C][Ring1][=Branch1][C][Ring1][Ring2]}&\\
    Can you give me the energy difference between the HOMO and LUMO orbitals of this molecule? 
    & 0.2756 \\
    \textit{SELFIES: [C][C][C][=Branch1][C][=O][N][Branch1][C][C][C][=Branch1][C][=O][N]}&\\
    What is the lowest unoccupied molecular orbital (LUMO) energy of this molecule?
    & -0.0064 \\
    \textit{SELFIES: [C][C][=C][O][C][=C][Ring1][Branch1][C][Branch1][C][C][C]}&\\
    Please provide the highest occupied molecular orbital (HOMO) energy of this molecule.
    & -0.2132 \\
    \bottomrule
    \end{tabular}
    }
    \label{tab:example_proprety_reg}
\end{table*}
\egroup

For regression, we adopt the instruction tuning data from \molins~\citep{mol-ins}. The regression based property prediction focuses on predicting the quantum mechanics properties of the molecules. 
The 1D sequence information in this task is given by SELFIES~\citep{selfies}.
The original data is sourced from the QM9 subset of the MolculeNet~\citep{molnet}. There are three subtasks: (i) Highest occupied molecular orbital (HOMO) energy prediction; (ii) Lowest occupied molecular orbital (LUMO) energy prediction; (iii) and HUMO-LUMO gap energy prediction. Some examples of the regression based property prediction dataset are given in Table~\ref{tab:example_proprety_reg}.

\subsection{Details of Reaction Prediction Dataset}
\label{appdx:reaction}

We adopt three chemical reaction related tasks from \molins~\citep{mol-ins}: Forward reaction prediction, reagent prediction, and retrosynthesis prediction. 
The input and output contain 1D sequence information given by SELFIES~\citep{selfies}. Some examples of the \molins datasets are given in Table~\ref{tab:example_reaction}, where the SELFIES represented molecules are denoted as ``<SELFIES>'' for clarity.

\bgroup
\def\arraystretch{1.2}
\begin{table*}[!ht]
    \centering
    \caption{
    \small
    Examples of the chemical reaction datasets.
    }
    \scalebox{0.75}{
    \begin{tabular}{l|p{14cm}}
    \toprule
    \textbf{Task} &\textbf{Examples} \\
    \midrule
    Forward Reaction Prediction& \textit{Question:} With the provided reactants and reagents, propose a potential product.\textit{<SELFIES>}\\
    & \textit{Answer:} \textit{<SELFIES>}\\
    Reagent Prediction& \textit{Question:} Please suggest some possible reagents that could have been used in the following chemical reaction. The reaction is \textit{<SELFIES>}\\
    & \textit{Answer:} \textit{<SELFIES>}\\
    Retrosynthesis Prediction& \textit{Question:} Please suggest potential reactants for the given product. The product is: \textit{<SELFIES>}\\
    & \textit{Answer:} \textit{<SELFIES>}\\
    \bottomrule
    \end{tabular}
    }
    \label{tab:example_reaction}
\end{table*}
\egroup

The task of forward reaction prediction aims to predict the possible products of a chemical reaction. The input includes the SELFIES sequences of the reactant and reagent of the chemical reaction. And the model needs to predict the SELFIES of the products.
The original data is sourced from USPTO~\footnote{\url{https://developer.uspto.gov/data}}, which consists of chemical reactions of organic molecules extracted from American patents and patent applications.

The task of reagent reaction prediction aims to predict the suitable catalysts, solvents, and ancillary substances with respect to a chemical reaction. The input includes the SELFIES sequences of the chemical reaction. 
The original data is sourced from USPTO~\footnote{\url{https://developer.uspto.gov/data}}, as the other tasks.

The task of retrosynthesis prediction aims to reverse engineer a particular compound by predicting the potential reactants or reagents that are required to synthesis the compound.
The input includes the SELFIES sequences of the target product. 
The original data is sourced from USPTO~\footnote{\url{https://developer.uspto.gov/data}}, similar to the other tasks.

\subsection{Details of Molecular Description Dataset}
\label{appdx:molcap}
For the molecular description task, we adopt a widely used dataset \chebi~\citep{edwards2021text2mol}. Based on the molecules from PubChem, \citet{edwards2021text2mol} collected the  Chemical Entities of Biological Interest (ChEBI)~\citep{chebi} annotations of the molecules, which are the descriptions of molecules. We transform the task into the instructions, and present some samples in Table~\ref{tab:example_molcap}.
The authors collect $33,010$ molecule-text pairs and split them into training ($80$\%), validation ($10$\%), and testing ($10$\%) subsets. We mainly adopt the original training split to tune the model and evaluate the tuned model on the original test split.

\bgroup
\def\arraystretch{1.2}
\begin{table*}[!ht]
    \centering
    \caption{
    \small
    Examples of the molecular descrioption datasets.
    }
    \scalebox{0.8}{
    \begin{tabular}{p{9cm}p{8cm}}
    \toprule
    \textbf{Question} &\textbf{Answer} \\
    \midrule
    \textit{SMILES: C1=CC=C(C=C1)[As](=O)(O)[O-]}& \\
    Could you give me a brief overview of this molecule? 
    & The molecule is the organoarsonic acid anion formed by loss of a single proton from the arsonic acid grouping in phenylarsonic acid. It is a conjugate base of a phenylarsonic acid. \\
    \textit{SMILES: CCCCCCCCCCCC(=O)OC(=O)CCCCCCCCCCC}& \\
    Could you provide a description of this molecule? 
    & The molecule is an acyclic carboxylic anhydride resulting from the formal condensation of the carboxy groups of two molecules of dodecanoic acid. It derives from a dodecanoic acid. \\
    \textit{SMILES: CCCCNC=O}& \\
    Please give me some details about this molecule. 
    & The molecule is a member of the class of formamides that is formamide substituted by a butyl group at the N atom. It has a role as a human metabolite. It derives from a formamide. \\
    \bottomrule
    \end{tabular}
    }
    \label{tab:example_molcap}
\end{table*}
\egroup

\subsection{Details of \hbench Dataset}
\label{appdx:hbench}
The \hbench is mainly used to measure the hallucination of common functional groups by LGLMs.
For the construction of \hbench, we consider the common functional groups in RDKit\footnote{\url{https://github.com/rdkit/rdkit/blob/master/Data/FunctionalGroups.txt}} as shown in Table~\ref{tab:rdkit_fg}. There are $39$ common functional groups, while we neglect the one with the name of ``???''.

Then, we leverage RDKit~\citep{rdkit} to detect the existence of the left $38$ valid functional groups within a molecule. We consider $3,300$ molecules from \chebi test split~\citep{edwards2021text2mol}, and adopt the query style as for large vision-language models~\citep{lvlm_hallu_eval} that queries the existence of specific functional group one by one:
\begin{myquotation}\centering
    \texttt{Is there a <functional group name> in the molecule?}
\end{myquotation}
Examples of \hbench are given in Table~\ref{tab:example_hbench}.

During the evaluation, we detect whether the LGLM gives outputs meaning ``Yes'' or ``No'' following the practice in~\citep{lvlm_hallu_eval}.
For each molecule, we construct questions with positive answers for all kinds of functional groups detected in the molecule, and questions with negative answers for randomly sampled $6$ functional groups from the $38$ common functional groups in RDKit.
The construction finally yields $23,924$ query answer pairs about the existence of functional groups in the molecule. While it is easy to scale up \hbench by automatically considering more molecules and a broader scope of functional groups, we find that the current scale is already sufficient to demonstrate the hallucination phenomena in LGLMs.

\bgroup
\def\arraystretch{1.2}
\begin{table*}[ht!]
    \centering
    \caption{
    \small
    List of functional groups from RDKit used to construct \hbench. The functional group with the name ``???'' is neglected.
    }
    \scalebox{0.8}{
    \begin{tabular}{ccc}
    \toprule
    \textbf{Chemical Representation} &\textbf{SMARTS}&\textbf{Name} \\
    \midrule
    -NC(=O)CH3&*-[N;D2]-[C;D3](=O)-[C;D1;H3]&methyl amide\\
    -C(=O)O&*-C(=O)[O;D1]&carboxylic acids\\
    -C(=O)OMe&*-C(=O)[O;D2]-[C;D1;H3]&carbonyl methyl ester\\
    -C(=O)H&*-C(=O)-[C;D1]&terminal aldehyde\\
    -C(=O)N&   *-C(=O)-[N;D1]&amide\\
    -C(=O)CH3&*-C(=O)-[C;D1;H3]&carbonyl methyl\\
    -N=C=O&   *-[N;D2]=[C;D2]=[O;D1]&isocyanate\\
    -N=C=S&   *-[N;D2]=[C;D2]=[S;D1]&isothiocyanate\\
    \midrule
    & \textit{Nitrogen containing groups}&\\
    \midrule
    -NO2&   *-[N;D3](=[O;D1])[O;D1]&nitro\\
-N=O&   *-[N;R0]=[O;D1]&nitroso\\
=N-O&   *=[N;R0]-[O;D1]&oximes\\
=NCH3&   *=[N;R0]-[C;D1;H3]&Imines\\
-N=CH2&   *-[N;R0]=[C;D1;H2]&Imines\\
-N=NCH3&   *-[N;D2]=[N;D2]-[C;D1;H3]&terminal azo\\
-N=N&   *-[N;D2]=[N;D1]&hydrazines\\
-N\#N&   *-[N;D2]\#[N;D1]&diazo\\
-C\#N&   *-[C;D2]\#[N;D1]&cyano\\
\midrule
    & \textit{S containing groups}&\\
    \midrule
    -SO2NH2&*-[S;D4](=[O;D1])(=[O;D1])-[N;D1]&primary sulfonamide\\
-NHSO2CH3&*-[N;D2]-[S;D4](=[O;D1])(=[O;D1])-[C;D1;H3]&methyl sulfonamide\\
-SO3H&*-[S;D4](=O)(=O)-[O;D1]&sulfonic acid\\
-SO3CH3&*-[S;D4](=O)(=O)-[O;D2]-[C;D1;H3]&methyl ester sulfonyl\\
-SO2CH3&*-[S;D4](=O)(=O)-[C;D1;H3]&methyl sulfonyl\\
-SO2Cl&*-[S;D4](=O)(=O)-[Cl]&sulfonyl chloride\\
-SOCH3&*-[S;D3](=O)-[C;D1]&methyl sulfinyl\\
-SCH3&*-[S;D2]-[C;D1;H3]&methylthio\\
-S&*-[S;D1]&thiols\\
=S&*=[S;D1]&thiocarbonyls\\
\midrule
    & \textit{Miscellaneous fragments}&\\
    \midrule
    -X&*-[\#9,\#17,\#35,\#53]&   halogens\\
-tBu&*-[C;D4]([C;D1])([C;D1])-[C;D1]&t-butyl\\
-CF3&*-[C;D4](F)(F)F&trifluoromethyl\\
-C\#CH&*-[C;D2]\#[C;D1;H]&acetylenes\\
-cPropyl&*-[C;D3]1-[C;D2]-[C;D2]1&cyclopropyl\\
\midrule
    & \textit{Teeny groups}&\\
    \midrule
-OEt&*-[O;D2]-[C;D2]-[C;D1;H3]&ethoxy\\
-OMe&*-[O;D2]-[C;D1;H3]&methoxy\\
-O&*-[O;D1]&side-chain hydroxyls\\
=O&*=[O;D1]&side-chain aldehydes or ketones\\
-N&   *-[N;D1]&primary amines\\
=N&   *=[N;D1]&???\\
\#N&   *\#[N;D1]&nitriles\\
    \bottomrule
    \end{tabular}
    }
    \label{tab:rdkit_fg}
\end{table*}
\egroup

\bgroup
\def\arraystretch{1.2}
\begin{table*}[!ht]
    \centering
    \caption{
    \small
    Examples of the \hbench dataset.
    }
    \scalebox{0.7}{
    \begin{tabular}{p{16cm}p{2cm}}
    \toprule
    \textbf{Question} &\textbf{Answer} \\
    \midrule
    \textit{SMILES: COC1=CC=CC2=C1C(=CN2)C/C(=N/OS(=O)(=O)[O-])/S[C@H]3[C@@H]([C@H]([C@@H]([C@H](O3)CO)O)O)O}&\\
    Is there a methyl ester sulfonyl group in the molecule? 
    & No \\
    \textit{SMILES: CN(C)C(=O)C(CCN1CCC(CC1)(C2=CC=C(C=C2)Cl)O)(C3=CC=CC=C3)C4=CC=CC=C4}&\\
    Is there a carbonyl methyl ester group in the molecule? 
    & Yes \\
    \textit{SMILES: CN(C)C(=O)C(CCN1CCC(CC1)(C2=CC=C(C=C2)Cl)O)(C3=CC=CC=C3)C4=CC=CC=C4}&\\
    Is there a terminal aldehyde group in the molecule? 
    & No \\
    \bottomrule
    \end{tabular}
    }
    \label{tab:example_hbench}
\end{table*}
\egroup

\section{Details of Experiments}
\label{appdx:exp}

\paragraph{Implementation of graph tokenizer.}
We implement the GNN tokenizer/encoder based on the same GNN backbone, which is a $5$-layer GIN~\citep{gin}. The hidden dimension is $300$. For the node-centric tokenization, we employ the VQVAE GNN tokenizer from \molebert~\citep{molebert} and adopt self-supervised learning tasks from the official \molebert implementation.\footnote{\url{https://github.com/junxia97/Mole-BERT}}
For \ours, we train the VQVAE with the self-supervised learning tasks from~\citep{zang2023hierarchical} based on the official implementation.\footnote{\url{https://github.com/ZangXuan/HiMol}}
Meanwhile, we set the hyperparameters of GNN tokenizer training the same as those recommended by~\citep{molebert,zang2023hierarchical}.

After training the tokenizer, we adopt the GNN encoder within the tokenizer instead of the codebook embeddings as we empirically find that the GNN embeddings perform better than that using the VQVAE codebook embeddings.

\paragraph{Implementation of LGLMs.}
For the cross-modal adapters, we implement it as a single-layer MLP with an input dimension of $300$ as our main focus is the tokenization. For \ours, we adopt three distinct adapters to handle the node-level, motif-level and graph-level embeddings. Meanwhile, we also adopt a Laplacian position encodings with respect to the supernode-augmented graphs. The dimension of the Laplacian position encoding is set to $8$, therefore the input dimensions of the adapters in \ours will be $308$.

For the LoRA adapters, we use a LoRA rank of $128$ and a scaling value $\alpha$ of $256$ for molecular property prediction (classification) in order to better fit with the task, and use a LoRA rank of $64$ and a scaling value $\alpha$ of $16$ for all the remaining methods and tasks.

For the base LLM, we mainly adopt \texttt{vicuna-v-1.3-7B}~\citep{vicuna}. The overall scale of parameters is around $6.9$B.

\paragraph{Implementation of instruction tuning.}
In stage 1 instruction tuning, we train all methods based on PubChem-$295$k dataset. 
The training goes $5$ epochs, with a batch size of $64$ (distributed to $4$ GPUs) by default.
If there is an OOM issue, we will decrease the batch size a little bit to $40$. The learning rate is set to $2\times 10^{-3}$ for all methods.

For classification-based property prediction, the training goes $20$ epochs, with a batch size of $128$ (distributed to $4$ GPUs) by default. If there is an OOM issue, we will decrease the batch size a little bit to $64$. 
The learning rate is set to $8\times 10^{-5}$ for all methods.

For regression-based property prediction, the training goes $5$ epochs, with a batch size of $64$ (distributed to $4$ GPUs) by default.
The learning rate is set to $2\times 10^{-5}$ for all methods.

For molecular description, the training goes $50$ epochs, with a batch size of $64$ (distributed to $4$ GPUs) by default. If there is an OOM issue, we will decrease the batch size a little bit to $32$. 
The learning rate is set to $8\times 10^{-5}$ for all methods.

For forward reaction prediction, the training goes $5$ epochs, with a batch size of $64$ (distributed to $4$ GPUs) by default.
The learning rate is set to $2\times 10^{-5}$ for all methods.

For reagent prediction, the training goes $5$ epochs, with a batch size of $64$ (distributed to $4$ GPUs) by default.
The learning rate is set to $2\times 10^{-5}$ for all methods.

For retrosynthesis prediction, the training goes $5$ epochs, with a batch size of $64$ (distributed to $4$ GPUs) by default.
The learning rate is set to $2\times 10^{-5}$ for all methods.

\paragraph{Training and evaluation.}
Throughout the paper, we use a max token length of $2048$.
Meanwhile, we adopt an AdamW optimizer with a warmup ratio of $3$\% for optimizing all models. We select the final model according to the best training loss.

For the evaluation of classification-based property prediction, we adopt the ROC-AUC following the common practice~\citep{molnet}. 

For the evaluation of regression-based property prediction, we adopt the Mean Absolute Error (MAE) following the common practice~\citep{mol-ins}. 

For the evaluation of molecular description, we adopt BLEU-2, BLEU-4, ROUGE-1, ROUGE-2, ROUGE-L, and METEOR following the common practice~\citep{papineni-etal-2002-bleu,lin-2004-rouge,edwards2021text2mol}. To improve the reliability of the evaluation, the metrics are computed based on the tokenizer \texttt{scibert\_scivocab\_uncased} of SciBERT~\citep{scibert}.

We follow the common practice to evaluate models for the tasks of chemical reaction predictions~\citep{mol-ins}.
We adopt linguistic metrics such as BLEU~\citep{papineni-etal-2002-bleu}, ROUGE-L~\citep{lin-2004-rouge},  METEOR~\citep{banerjee-lavie-2005-meteor} and Levenshtein scores~\citep{levenshtein}. Meanwhile, we also validate the validity of the generated molecular sequences with RDKit~\citep{rdkit}. In addition, several molecular similarity measures are also leveraged. Specifically, we present the MAE of the RDKit, MACCS, and Morgan fingerprints to assess the semantic similarity of the generated compounds and the ground truth ones~\citep{Durant2002ReoptimizationOM,Schneider2015GetYA}.

As for the \hbench, in order to avoid the drawbacks that LGLMs may output answers that do not follow the instructions, we compare the loss values by feeding the answers of ``Yes'' and ``No'', and take the one with a lower autoregressive language modeling loss as the answer. Following the practice in LVLMs, we present the F1 scores, accuracies, and the ratio that the model answers ``Yes''~\citep{lvlm_hallu_eval}. Given the severe imbalance of positive and negative samples, we separately report the F1 scores for positive and negative classes.

\paragraph{Software and hardware.}
We implement our methods with PyTorch 11.3~\citep{pytorch}.
We run experiments on Linux Servers with NVIDIA V100 and NVIDIA A100 (40G) graphics cards with CUDA 11.7.

\section{More Ablation Studies}
\label{appdx:ablation}

\subsection{Computation overhead}
\label{appdx:computation}

\begin{table}[ht]
\label{tab:pretraining_time}
\caption{Training Computational Overhead. We count the average graph size of PubChem and HiPubChem, where HiPubChem adds 9 additional tokens on average. The real preprocessing time and training time are shown below, which are estimated based on 4 A100 40G GPUs. Although HIGHT requires more time to train, the absolute computational overhead of HIGHT is not high.}
\centering
\begin{tabular}{lccc}
\toprule
                                 & Graph Size & Preprocessing Time & Training Time      \\\midrule
{ PubChem}   & { 34.39}                                                  & { 16min 32sec}        & { 8hour 17min 59sec}  \\
{ HiPubChem} & { 43.21}                                                  & { 25min 35sec}        & { 15hour 36min 23sec}\\\bottomrule
\end{tabular}
\end{table}

\begin{table}[ht]
\label{tab:inference_time}
\caption{ Inference Computational Overhead. The inference computational overhead is estimated based on 4 A100 40G GPUs. During the inference, the LLM latency takes up the majority of time. A well-trained LGLM with HIGHT is able to generate more concise and valid answers and thus may take less time during inference.}
\centering
\resizebox{\textwidth}{!}{
\begin{tabular}{
lccccc}\toprule
                                   & { \textbf{Property Prediction}} & { \textbf{MolCaption}} & { \textbf{Reagent Prediction}} & { \textbf{Forward Reaction}} & { \textbf{Retrosynthesis}} \\\midrule
{ InstructMol} & { 14min 54sec}                  & { 6hour 22min 27sec}   & { 56min 56sec}                 & { 1hour 34min 28sec}         & { 1hour 50min 47sec}       \\
{ HIGHT}       & { 15min 12sec}                  & { 4hour 59min 50sec}   & { 50min 29sec}                 & { 1hour 22min 08sec}         & { 1hour 49min 42sec}      \\\bottomrule
\end{tabular}}
\end{table}

\begin{table}[ht]
\label{tab:param_count}
\caption{Number of Tunable Parameters during Training. When pretraining the GNN tokenizer, the number of tunable parameters is the number of parameters in GNN encoder; In stage 1, the number of tunable parameters is the number of parameters in the projector; In stage 2, the number of tunable parameters is the number of parameters in the projector and in LoRA.
}
\centering
\resizebox{\textwidth}{!}{
\begin{tabular}{
lccccc}\toprule
                                   & { \textbf{graph token dimension}} & \multicolumn{1}{l}{{ \textbf{GNN encoder}}} & \multicolumn{1}{l}{{ \textbf{params in projector}}} & \multicolumn{1}{l}{{ \textbf{params in tokenizer}}} & \multicolumn{1}{l}{{ \textbf{LoRA}}} \\\midrule
{ InstructMol} & { 300d}                           & { 1,860,905}                                                                         & { 1,232,896}                                                                       & { 3,093,801}                                                                       & { 159,907,840}                                               \\
{ HIGHT}       & { 300d}                           & { 1,865,105}                                                                         & { 3,796,992}                                                                       & { 5,662,097}                                                                       & { 159,907,840}                                              \\\bottomrule
\end{tabular}}
\end{table}

\subsection{Ablation studies with different setups of the tokenizers}
\label{appdx:ablation_tokenizer}
In Table~\ref{tab:chemical_reaction_ablation}, we present more results of the ablation studies with different setups of \ours and node-centric tokenizer.

\begin{table*}[h!]
\centering
\small
\vspace{-0.1in}
\caption{\footnotesize
{
More results of chemical reaction tasks with ablation studies. These tasks encompass reagent prediction, forward reaction prediction, and retrosynthesis. $\dagger$: few-shot ICL results from \cite{mol-ins}. $*$: use task-specific instruction data to finetune.}
}
\label{tab:chemical_reaction_ablation}
\resizebox{\textwidth}{!}{
\begin{tabular}{lccccccc}
\toprule
\textsc{Model}
&\textsc{Exact}$\uparrow$  & \textsc{BLEU}$\uparrow$  & \textsc{Levenshtein}$\downarrow$  & \textsc{RDK FTS}$\uparrow$  & \textsc{MACCS FTS}$\uparrow$ & \textsc{Morgan FTS}$\uparrow$ & \textsc{Validity}$\uparrow$ \\

\midrule[1.1pt]
\rowcolor[rgb]{0.94, 0.97, 1.0}
\multicolumn{8}{l}{\textit{Reagent Prediction}} \\
Alpaca$^\dagger$~\citep{alpaca} & 0.000 & 0.026 & 29.037 & 0.029 & 0.016 & 0.001 & 0.186 \\
Baize$^\dagger$~\citep{baize} & 0.000 & 0.051 & 30.628 & 0.022 & 0.018 & 0.004 & 0.099 \\
ChatGLM$^\dagger$~\citep{zeng2023glmb} & 0.000 & 0.019 & 29.169 & 0.017 & 0.006 & 0.002 & 0.074 \\
LLama$^\dagger$~\citep{llama} & 0.000 & 0.003 & 28.040 & 0.037 & 0.001 & 0.001 & 0.001 \\
Vicuna$^\dagger$~\citep{vicuna} & 0.000 & 0.010 & 27.948 & 0.038 & 0.002 & 0.001 & 0.007 \\
Mol-Instruction~\citep{mol-ins} & 0.044 & 0.224 & \textbf{23.167} & 0.237 & \textbf{0.364 }& 0.213 & 1.000 \\
LLama-7b$^*$~\citep{llama}(LoRA) &0.000	&0.283	&53.510	&0.136	&0.294	&0.106	&1.000 \\
{InstructMol-G}       & 0.031	&0.429	&31.447	&0.389	&0.249	&0.220	&1.000\\
\ {+Positional Encoding}   & 0.009	&0.423	&30.833	&0.370	&0.231 &0.197	&0.986 \\
\ {+\hdata}       & 0.016	&0.473	&30.455	&0.369	&0.237	&0.194	&0.990 \\
\ {+Large Tokenizer}       & 0.040	&0.454	&29.163	&0.416	&0.284	&0.248	&1.000 \\
{InstructMol-GS} & 0.057	&0.439	&29.757 &0.437	&0.314	&0.271	&0.999\\
{InstructMol+LLama-2-7b-chat} & 0.016	&0.459	&29.238 &0.359	&0.225	&0.189	&0.988\\
\midrule
\textbf{\ourst-G}       & 0.050	&0.462	&28.970	&0.441	&0.314	&0.275	&1.000\\
{\textbf{\ourst-GS}}       & \textbf{0.067}	&0.482	&27.167	&\textbf{0.462}	&0.346	&\textbf{0.303}	&1.000\\
{\textbf{\ourst+LLama-2-7b-chat}}       & 0.057	&\textbf{0.495}	&{26.591}	&0.453	&0.333	&0.293	&1.000\\
\midrule[1.1pt]
\rowcolor[rgb]{0.94, 0.97, 1.0}
\multicolumn{8}{l}{\textit{Forward Reaction Prediction}} \\
Alpaca$^\dagger$~\citep{alpaca} & 0.000 & 0.065 & 41.989 & 0.004 & 0.024 & 0.008 & 0.138 \\
Baize$^\dagger$~\citep{baize} & 0.000 & 0.044 & 41.500 & 0.004 & 0.025 & 0.009 & 0.097 \\
ChatGLM$^\dagger$~\citep{zeng2023glmb} & 0.000 & 0.183 & 40.008 & 0.050 & 0.100 & 0.044 & 0.108 \\
LLama$^\dagger$~\citep{llama} & 0.000 & 0.020 & 42.002 & 0.001 & 0.002 & 0.001 & 0.039 \\
Vicuna$^\dagger$~\citep{vicuna} & 0.000 & 0.057 & 41.690 & 0.007 & 0.016 & 0.006 & 0.059 \\
Mol-Instruction~\citep{mol-ins} & 0.045 & 0.654 & 27.262 & 0.313 & 0.509 & 0.262 & 1.000 \\
LLama-7b$^*$~\citep{llama}(LoRA) &0.012	&0.804	&29.947	&0.499	&\textbf{0.649}	&{0.407}	&1.000 \\
{InstructMol-G}        & 0.031 &0.853 &24.790	&0.512	&0.362	&0.303	&0.993 \\
\ {+Positional Encoding}   & 0.0102	&0.829	&26.622	&0.419	&0.328	&0.268	&0.981 \\
\ {+\hdata}       & 0.011	&0.819	&26.010	&0.396	&0.315	&0.264	&0.975 \\
\ {+Large Tokenizer}       & 0.040	&0.861	&24.051	&0.544 &0.380	&0.328	&0.996 \\
{InstructMol-GS} & {0.252} &{0.926}	&{17.773}	&{0.755}	&{0.599}	&{0.543}	&1.000\\
{InstructMol+LLama-2-7b-chat} & 0.020	&0.841	&25.109 &0.426	&0.339	&0.284	&0.998\\
\midrule
\textbf{\ourst-G}        & 0.037 &0.869 &{23.759}	&{0.590}	&0.394	&0.340	&0.993 \\
{\textbf{\ourst-GS}} & \textbf{0.293} &\textbf{0.935}	&\textbf{16.687}	&\textbf{{0.774}}	&{0.618}	&\textbf{{0.566}}	&1.000\\
{\textbf{\ourst+LLama-2-7b-chat}}       & 0.042	&0.873	&{23.854}	&0.590	&0.402	&0.344	&0.996\\
\midrule[1.1pt]
\rowcolor[rgb]{0.94, 0.97, 1.0}
\multicolumn{8}{l}{\textit{Retrosynthesis}} \\
Alpaca$^\dagger$~\citep{alpaca} & 0.000 & 0.063 & 46.915 & 0.005 & 0.023 & 0.007 & 0.160 \\
Baize$^\dagger$~\citep{baize} & 0.000 & 0.095 & 44.714 & 0.025 & 0.050 & 0.023 & 0.112 \\
ChatGLM$^\dagger$~\citep{zeng2023glmb} & 0.000 & 0.117 & 48.365 & 0.056 & 0.075 & 0.043 & 0.046 \\
LLama$^\dagger$~\citep{llama} & 0.000 & 0.036 & 46.844 & 0.018 & 0.029 & 0.017 & 0.010 \\
Vicuna$^\dagger$~\citep{vicuna} & 0.000 & 0.057 & 46.877 & 0.025 & 0.030 & 0.021 & 0.017 \\
Mol-Instruction~\citep{mol-ins} & 0.009 & 0.705 & 31.227 & 0.283 & 0.487 & 0.230 & 1.000 \\
LLama-7b$^*$~\citep{llama}(LoRA) &0.000	&0.283	&53.510	&0.136	&0.294	&0.106	&1.000 \\
{InstructMol-G}       & 0.001	&0.835	&31.359	&0.447	&0.277	&0.241	&0.996\\
\ {+Positional Encoding}   & 0.000	&0.793	&33.859	&0.295	&0.218	&0.192	&0.983 \\
\ {+\hdata}       & 0.000	&0.755	&35.811	&0.282	&0.218	&0.177	&0.997 \\
\ {+Large Tokenizer}       & 0.001	&0.842	&30.613	&0.459	&0.287	&0.263	&0.999 \\
{InstructMol-GS} & 0.172	&0.911	&20.300 &0.765	&0.615	&0.568	&1.000\\
{InstructMol+LLama-2-7b-chat} & 0.000	&0.806	&32.128 &0.292	&0.234	&0.202	&0.985\\
\midrule
\textbf{\ourst-G}       & 0.008	&0.863	&{28.912}	&0.564	&0.340	&0.309	&1.000\\
{\textbf{\ourst-GS}}       & \textbf{0.202}	&\textbf{0.914}	&\textbf{20.194	}&\textbf{0.772}	&\textbf{0.623}	&\textbf{0.577}	&0.999\\
{\textbf{\ourst+LLama-2-7b-chat}}       & 0.006	&0.865	&{28.964}	&0.563	&0.338	&0.306	&0.999\\
\bottomrule
\end{tabular}
}
\vspace{-0.1in}
\end{table*}

\begin{table*}
\centering
\small
\caption{\footnotesize
{
Full results of motif hallucinations on \hbench with ablation studies.
}}
\label{tab:hbench_ablation}
\scalebox{1}{
\begin{tabular}{lccc}
    \toprule
    \textsc{Method}  &F1 (pos) $\uparrow$  &F1 (neg) $\uparrow$ &F1 (avg) $\uparrow$ \\
    \midrule
    \rowcolor[rgb]{0.94, 0.97, 1.0} \multicolumn{4}{l}{\textit{Node-centric Tokenization}} \\
    {InstructMol-G} & 95.7 & 9.5 &52.6\\
    {{InstructMol-G}+LLama-2-7b-chat} & \textbf{99.6} & 2.8 &51.2\\
    {InstructMol-GS} & 97.1 & 10.6 &53.8\\
    \midrule
    \rowcolor[rgb]{0.94, 0.97, 1.0} \multicolumn{4}{l}{\textit{Hierarchical Tokenization}} \\
    \textbf{\ourst-G}& {85.5} & 48.2 & 66.9\\
    {\textbf{\ourst-G}+LLama-2-7b-chat}& {55.1} & \textbf{65.2}& \textbf{60.2} \\
    \textbf{\ourst-GS} & {84.5} & 42.7 &63.6\\
    \midrule
    \rowcolor[rgb]{0.94, 0.97, 1.0} \multicolumn{4}{l}{\textit{Ablation variants}} \\
    {InstructMol-G + Positional Encoding}& 96.4 & 19.8 &58.1\\
    {InstructMol-G + \hdata}& 96.6 & 12.5&54.6 \\
    \ourst-G w/o \hdata& 96.6 & 12.5&54.6\\
    \ourst-GS w/o \hdata& 98.2 & 6.5&52.4\\
    \bottomrule
\end{tabular}
}
\end{table*}

\begin{table*}
\centering
\small
\caption{\footnotesize
{
Results of molecular property prediction tasks (regression) on QM9 with ablation studies. We report the result in MAE. $\dagger$: few-shot in-context learning (ICL) results from~\citep{mol-ins}. $\Delta{\epsilon}$ refers to the HOMO-LUMO energy gap.}
}
\label{tab:property_pred_ablation}
\scalebox{1}{
    \begin{tabular}{lccccc}
        \toprule
        \textsc{Method} &\textsc{HOMO} $\downarrow$ &\textsc{LUMO} $\downarrow$  &$\Delta{\epsilon}$ $\downarrow$ &\textsc{Avg} $\downarrow$\\
        \midrule 
        \rowcolor[rgb]{0.94, 0.97, 1.0}
        \multicolumn{5}{l}{\textit{LLM Based Generalist Models}} \\
        Alpaca$^\dagger$~\citep{alpaca} & - & - & - & 322.109\\
        Baize$^\dagger$~\citep{baize} & - & - & - & 261.343\\
        LLama2-7B~\citep{llama2} (5-shot ICL) & 0.7367 & 0.8641 & 0.5152 & 0.7510 \\
        Vicuna-13B~\citep{vicuna} (5-shot ICL) & 0.7135 & 3.6807 & 1.5407 & 1.9783 \\
        Mol-Instruction~\citep{mol-ins} & 0.0210 & 0.0210 & 0.0203 & 0.0210 \\
        {InstructMol-G} &  0.0111 & 0.0133 & 0.0147 & 0.0130 \\
        {\ +Positional Encodings} &  0.0300 & 0.0395 & 0.0357 & 0.0350 \\
        {\ +\hdata} &  0.0305 & 4.4019 & 0.0494 & 1.1226 \\
        \midrule
        \textbf{\ourst-G} &  \textbf{0.0078} & \textbf{0.0086} & \textbf{0.0095} & \textbf{0.0086} \\
        \bottomrule
    \end{tabular}
}
\end{table*}

\end{document}